\pdfoutput=1
\def\arxivpreprint{}
\documentclass{article} % For LaTeX2e
\usepackage{iclr2027_conference,times}

\usepackage{amsmath,amsfonts,bm}

\def\eqref#1{equation~\ref{#1}}
\def\1{\bm{1}}

\DeclareMathAlphabet{\mathsfit}{\encodingdefault}{\sfdefault}{m}{sl}
\SetMathAlphabet{\mathsfit}{bold}{\encodingdefault}{\sfdefault}{bx}{n}

\usepackage[T1]{fontenc}
\usepackage{hyperref}
\ifdefined\arxivpreprint
\hypersetup{%
  pdftitle={FuseAlign: Forced Alignment in the Wild},
  pdfauthor={Mithilesh Vaidya, Stephen Bailey, Sumukh Badam, Matthew Bendel, Xingzhe He}%
}
\fi
\usepackage{url}
\usepackage{amsmath, amssymb}
\usepackage{booktabs}
\usepackage{multirow}
\usepackage{graphicx}
\usepackage{xcolor}
\usepackage{enumitem}

\definecolor{improve}{RGB}{0,128,0}
\definecolor{degrade}{RGB}{200,0,0}
\newcommand{\better}[1]{\textcolor{improve}{#1}}
\newcommand{\worse}[1]{\textcolor{degrade}{#1}}
\newcommand{\ci}[1]{{\scriptsize$\pm$#1}}
\newcommand{\sys}{FuseAlign}
\newcommand{\bench}{AlignBench}
\newcommand{\scribe}{Scribe}
\newcommand{\whisper}{Whisper}
\newcommand{\nullint}{(\varnothing_s,\varnothing_e)}

\newcommand{\papertitle}{\sys{}: Forced Alignment in the Wild}
\title{\papertitle}

\ifdefined\arxivpreprint
\usepackage{tgheros}
\usepackage[most]{tcolorbox}
\definecolor{brandrecred}{HTML}{F73B3B}
\definecolor{brandlogotype}{HTML}{651A39}
\definecolor{cardfg}{HTML}{26171D}
\definecolor{cardbg}{HTML}{FFF8F5}
\definecolor{cardrule}{HTML}{EBDFDA}
\definecolor{cardshadow}{HTML}{DDD3CF}
\newcommand{\cardname}[1]{\mbox{\textbf{#1}}}
\hypersetup{colorlinks=true, linkcolor=brandlogotype, citecolor=brandlogotype, urlcolor=blue!70!black}
\fi

\author{
\begin{tabular*}{\dimexpr\textwidth-2\tabcolsep-3pt\relax}{@{\extracolsep{\fill}}ccc@{}}
\noalign{\vskip 40pt}
Mithilesh Vaidya & Stephen Bailey & Sumukh Badam \\
{\normalfont Descript} & {\normalfont Descript} & {\normalfont Descript} \\
{\scriptsize\normalfont\texttt{mithilesh@descript.com}} &
{\scriptsize\normalfont\texttt{stephen@descript.com}} &
{\scriptsize\normalfont\texttt{sumukh@descript.com}} \\[10pt]
\multicolumn{3}{c}{
\begin{tabular}{@{}c@{\hspace{8em}}c@{}}
Matthew Bendel & Xingzhe He \\
{\normalfont Descript} & {\normalfont Descript} \\
{\scriptsize\normalfont\texttt{matt.bendel@descript.com}} &
{\scriptsize\normalfont\texttt{xingzhe@descript.com}}
\end{tabular}}
\end{tabular*}
}

\ifdefined\arxivpreprint
\iclrfinalcopy
\fi

\begin{document}

\ifdefined\arxivpreprint
% Branded title card (arXiv only). The ICLR build uses the style's \maketitle
% and abstract environment below.
\lhead{Preprint. Under review.}
\begin{tcolorbox}[enhanced, colback=cardbg, colframe=cardrule, boxrule=0.4pt,
  arc=8pt, left=12pt, right=12pt, top=14pt, bottom=12pt, boxsep=0pt,
  drop fuzzy shadow=cardshadow]
{\sffamily\bfseries\fontsize{19}{23}\selectfont\color{cardfg}\raggedright
\papertitle\par}
\vspace{7pt}
{\sffamily\large\color{cardfg}\raggedright\hyphenpenalty=10000\exhyphenpenalty=10000
\cardname{Mithilesh Vaidya}, \cardname{Stephen Bailey}, \cardname{Sumukh Badam},
\cardname{Matthew Bendel}, \cardname{Xingzhe He}\par}
\vspace{4pt}
{\color{cardfg}Descript, Inc.\par}
{\small\texttt{\{mithilesh, stephen, sumukh, matt.bendel, xingzhe\}@descript.com}\par}
\vspace{11pt}
{\color{cardfg}Word-level forced alignment estimates when each transcript word occurs in an audio recording. It underpins text-based media editing, subtitling, speech-data curation, and phonetic analysis. Existing evaluations understate the difficulty of forced alignment by relying on short, clean speech, perfect transcripts, and metrics that obscure consequential alignment errors. In contrast, real-world media and data-processing pipelines operate on long and diverse recordings. Additionally, forced aligners often operate on error-prone automatic speech recognition (ASR) output. We address these gaps with improved evaluation metrics, a scoring protocol for real ASR transcripts, and \bench{}, a benchmark spanning diverse speaker, acoustic, and text conditions. We further introduce \sys{}, a transformer-based aligner trained on large-scale pseudo-labeled speech with online label correction. \sys{} performs joint contextualization of audio and text for the localization of coarse words. The model then refines boundaries at millisecond resolution and detects missing transcript words in the audio without lexicon-based or Viterbi decoding. On \bench{}, \sys{} substantially outperforms all baselines and remains robust under real ASR transcripts. Ablations show that convolutional upsampling and EMA-snapshot label correction matter more than model properties such as parameter count.
\par}
\vspace{9pt}
\noindent
\begin{minipage}[b]{0.72\textwidth}
% Uncomment once the benchmark and code are public.
% {\sffamily\bfseries\small Project Page:} {\small\url{https://descriptinc.github.io/fusealign/}}\\
% {\sffamily\bfseries\small Code:} {\small\url{https://github.com/descriptinc/fusealign}}
\end{minipage}%
\hfill
\raisebox{0pt}[0pt][0pt]{\includegraphics[height=13pt]{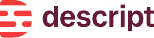}}
\end{tcolorbox}
\vspace{6pt}
\else
\maketitle
\fi

\ifdefined\arxivpreprint\else
\begin{abstract}
Word-level forced alignment estimates when each transcript word occurs in an audio recording. It underpins text-based media editing, subtitling, speech-data curation, and phonetic analysis. Existing evaluations understate the difficulty of forced alignment by relying on short, clean speech, perfect transcripts, and metrics that obscure consequential alignment errors. In contrast, real-world media and data-processing pipelines operate on long and diverse recordings. Additionally, forced aligners often operate on error-prone automatic speech recognition (ASR) output. We address these gaps with improved evaluation metrics, a scoring protocol for real ASR transcripts, and \bench{}, a benchmark spanning diverse speaker, acoustic, and text conditions. We further introduce \sys{}, a transformer-based aligner trained on large-scale pseudo-labeled speech with online label correction. \sys{} performs joint contextualization of audio and text for the localization of coarse words. The model then refines boundaries at millisecond resolution and detects missing transcript words in the audio without lexicon-based or Viterbi decoding. On \bench{}, \sys{} substantially outperforms all baselines and remains robust under real ASR transcripts. Ablations show that convolutional upsampling and EMA-snapshot label correction matter more than model properties such as parameter count.

\end{abstract}
\fi

\section{Introduction}
\label{sec:intro}

Word-level forced alignment refers to the process of estimating timestamps for every word in a transcript given an audio recording. This primitive drives text-based editing \citep{fried2019text,tan2021editspeech}, subtitles \citep{papi2023direct,rastorgueva2023nemo}, speech-data preparation \citep{panayotov2015librispeech,GigaSpeech2021,kurzinger2020ctc,yang2025gigaspeech,peng25c_interspeech}, and linguistic analysis \citep{mcauliffe2017montreal,ahn2022voxcommunis,chodroff2024phonetic}. These applications emphasize different aspects of alignment quality. Text-based editing needs fine-grained temporal control: errors of tens of milliseconds can clip part of a consonant, and errors of a few hundred milliseconds can move an edit boundary into a neighboring word. However, text-based editing is largely insensitive to where a boundary falls 
within adjacent silence because moving a cut within a pause does not alter audible 
speech. Subtitling can tolerate a few hundred milliseconds at the boundary but requires robustness to multiple speakers and overlapping speech. In data curation, large-scale corpora are built by aligning long recordings to often noisy text,  which is obtained by automatic speech recognition (ASR) models. A few gross misalignments can contaminate the resulting corpora in such cases. These differences call for evaluation metrics that reflect application-specific failure modes and benchmarks that resemble real deployment conditions. Existing forced-alignment evaluations fall short on both fronts.

First, existing benchmarks do not represent real-world recordings. The recent MFA~3.0 evaluation includes spontaneous English, Japanese, and Korean, but still uses curated utterances whose transcripts match the audio \citep{mcauliffe2026montreal}. Popular English corpora are also narrow. TIMIT contains short, read speech \citep{garofolo1993timit}, while Buckeye contains laboratory-recorded conversations \citep{pitt2007buckeye}. Both include only American speakers. Articulatory corpora such as MOCHA-TIMIT, mngu0, USC-TIMIT, and the Wisconsin XRMB also provide manual timing and are valuable for studying speech articulation, but they contain scripted prompts read by one or a few speakers \citep{wrench2000mocha,richmond2011announcing,narayanan2014real,westbury1994x}. None of these corpora captures the noisy, reverberant, long-form, multi-speaker, and accent-diverse conditions common in media pipelines. Large in-the-wild ASR corpora such as GigaSpeech, People's Speech, and VoxPopuli do cover these conditions \citep{GigaSpeech2021,galvez2021peoplesspeech,wang2021voxpopuli}, but lack hand-corrected word boundaries.

Second, conventional metrics obscure consequential alignment failures. Absolute boundary error penalizes a boundary that lands harmlessly inside a pause exactly as much as one that cuts into neighboring speech. This measures timestamp disagreement rather than what a user hears when editing media. Similarly, averaging a word's start and end error can hide a large error at either edge, even though a single misplaced boundary is sufficient to clip the word or neighboring speech. Finally, evaluations that score only successfully aligned words can make systems that reject difficult words appear artificially accurate~\citep{rousso2024tradition}.

Third, all popular benchmarks align hand-corrected transcripts that agree with the audio. In contrast, media and data-curation pipelines often align ASR output containing substituted, dropped, and hallucinated words. MFA notes that transcript and pronunciation deviations can occasionally cause catastrophic misalignment, but does not systematically evaluate this setting \citep{mcauliffe2026montreal}. 
% Section~\ref{sec:transcript-errors} describes how we score aligners on ASR transcripts, and we report results on ElevenLabs Scribe v2 \citep{elevenlabs2026scribe} and Whisper large-v3 \citep{radford2023robust}.
Real ASR systems also vary substantially in transcription quality. In our experiments, ElevenLabs Scribe v2 \citep{elevenlabs2026scribe} and Whisper large-v3 \citep{radford2023robust} provide two representative operating points.

These evaluation gaps reflect a broader modeling challenge: existing aligner families trade fine boundary precision against robustness to transcript and pronunciation variation. Lexicon-and-HMM systems such as Gentle, MFA, and Nyra rely on pronunciation resources and globally constrained Viterbi decoding, so a local transcript or pronunciation mismatch can propagate through the alignment \citep{ochshorn2016gentle,mcauliffe2017montreal,mcauliffe2026montreal,nyralabs2026forcedaligner}. CTC-based aligners, which fit the transcript's characters to the frame-level posteriors of a CTC-trained ASR model \citep{bain2022whisperx,rastorgueva2023nemo,pratap2024scaling}, trade precision for robustness: broad training and character vocabularies reduce their dependence on pronunciation lexicons, but their objectives do not supervise precise word boundaries, and they often lag behind dedicated aligners at fine tolerances. Speech-LLM aligners share similar vocabulary flexibility but predict quantized timestamps at resolutions too coarse for precise editing. For example, \citet{qwen3asr2026} predicts timestamps on an 80\,ms grid, far coarser than the 10\,ms grid of aligners like Gentle. These limitations suggest that a robust aligner should combine global audio-text context with fine-grained boundary prediction, while avoiding a single globally constrained decoding path through which local transcript mismatches can propagate.

To address these evaluation and modeling gaps, we contribute: (i) a revised evaluation framework for hand-corrected and real ASR transcripts that captures real-world usage of forced aligners, (ii) \bench{}, a hand-corrected benchmark spanning diverse speaker, acoustic, and text conditions, and (iii) \sys{}, a transformer-based forced aligner that jointly represents words and audio to predict word presence and precise word boundaries at fine temporal resolution. We show that \sys{} consistently outperforms other aligners and retains strong performance on ASR transcripts.

\section{Related Work}
\label{sec:related}

\textbf{Lexicon-and-HMM aligners.}
Classical systems, from P2FA, FAVE, Prosodylab-Aligner, and WebMAUS \citep{yuan2008speaker,rosenfelder2011fave,gorman2011prosodylab,kisler2012signal} to their Kaldi-based successors MFA and Gentle, expand the transcript through a pronunciation lexicon into a phone-level HMM graph, score frames with an acoustic model, and select one global Viterbi path \citep{povey2011kaldi}. The are precise: MFA~3.0 reports mean word-boundary errors of 19-25\,ms on TIMIT and Buckeye \citep{mcauliffe2026montreal}. Gentle re-decodes unmatched regions and rejects words it cannot locate; MFA adds acoustic-model adaptation, G2P, and cross-language phone remapping but expects a matching transcript. Nyra keeps this GMM-HMM recipe with frozen WavLM-Large features, a noise-robust LDA projection, and dedicated phones for fillers and non-lexical vocal events.

\textbf{CTC-based aligners.}
Aligners such as ctc-segmentation, NeMo NFA, and MMS-FA fit the transcript's characters to the frame-level posteriors of a CTC-trained ASR model with a transcript-constrained Viterbi search \citep{kurzinger2020ctc}; WhisperX does the same with an English letter-level wav2vec\,2.0 model. Broad training and flexible vocabularies make them robust to lexical and acoustic variation, but their objective does not directly identify word edges \citep{rousso2024tradition}. MWA adds self-supervised phone-boundary features and learned dynamic programming at 10\,ms resolution, improving on MFA and MMS, but still assumes a matching transcript and one global path \citep{weber2026multilingual}.

\textbf{Whisper and Speech-LLM aligners.}
Whisper-based aligners read word times from decoder cross-attention over their own hypothesis. CrisperWhisper~2.0 selects the ten decoder heads whose untrained cross-attention already correlates with word location and supervises only those, matching their average to MFA occupancy targets with a word-level cosine loss, then reading times from a Viterbi path over the pooled attention \citep{wagner2026transcription}. It times the words it emits rather than a supplied transcript (Appendix~\ref{app:related}). Speech-LLM aligners instead predict discrete timestamp tokens \citep{mu2026llm,qwen3asr2026}; these support flexible vocabularies but are too coarse for precise editing. \sys{} directly and independently supervises each word boundary. Appendix~\ref{app:related} covers additional systems and implementation details.

\textbf{Current evaluation practice.}
\label{sec:current-eval}
Forced aligners are evaluated against hand-labeled word boundaries on curated corpora whose transcripts match the audio, such as TIMIT and Buckeye. Studies report the mean boundary error and the fraction of boundaries within fixed tolerances, often only over the words an aligner places \citep{rousso2024tradition,mcauliffe2026montreal}. This protocol suits analyses of phone and word durations, but it leaves open whether an error cuts audible speech, whether both edges of a word are usable, and how rejected words and ASR errors affect the result. Section~\ref{sec:metrics} defines these metrics, examines their shortcomings, and extends them to answer these questions.

\section{An Evaluation Framework for Word-Level Alignment}
\label{sec:metrics}

\textbf{Task and notation.} An aligner receives a waveform and a transcript of $N$ words, $(w_1, \dots, w_N)$. For each word, it returns an interval $(\hat{s}_i,\hat{e}_i)$ or the null interval $\nullint$ if the word cannot be located. The reference gives each word an interval $(s_i,e_i)$. For transcripts that match the reference, we define each conventional metric, discuss its shortcomings for editing, and introduce our modification; we then extend the protocol to ASR transcripts. Appendix~\ref{app:metrics} gives precise definitions and edge cases.

\begin{figure}[t]
\centering
\includegraphics[width=\textwidth]{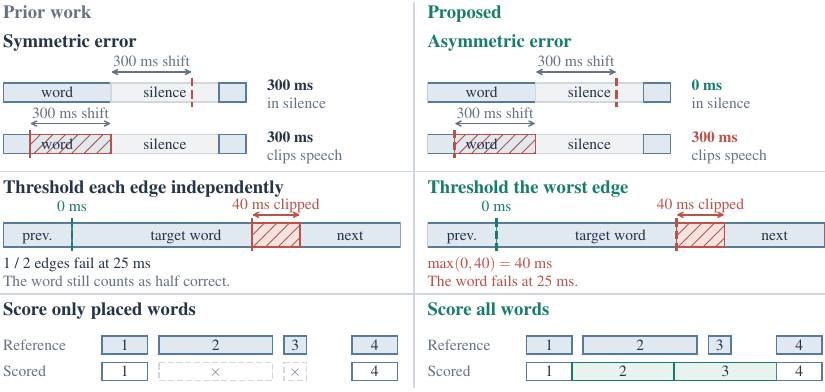}
\caption{\textbf{Prior versus proposed evaluation of word-level alignment.} Each row is a failure mode. Top: symmetric error penalizes a 300\,ms shift equally whether it stays in silence or clips speech; asymmetric error penalizes only the latter. Middle: thresholding each edge independently counts a word with one 40\,ms clip as half correct; thresholding the worst edge fails the word. Bottom: scoring only placed words conceals rejected words; a uniform fallback scores every transcript word.}
\label{fig:eval-metrics}
\end{figure}

\textbf{Boundary error.}
\label{sec:boundary}
The standard measure of alignment accuracy is the mean absolute error (MAE) between predicted and reference boundaries, averaged over all $2N$ start and end boundaries \citep{mcauliffe2017montreal}:
\[
\mathrm{MAE}=\frac{1}{2N}\sum_{i=1}^{N}\bigl(|\hat{s}_i-s_i|+|\hat{e}_i-e_i|\bigr).
\]
We introduce the term \emph{symmetric error} for the per-boundary displacement $|\hat{s}_i-s_i|$ or $|\hat{e}_i-e_i|$, since it penalizes early and late predictions, and silence and speech, alike. However, for many real-world applications, such as editing, a boundary that moves into the silence next to a word changes no audible speech, whereas the same movement into speech clips it (Figure~\ref{fig:eval-metrics}, top). We therefore propose \emph{asymmetric error}, which penalizes only the latter. Each boundary may fall anywhere in a free interval spanning the adjacent pause, and the error, $\varepsilon^{\mathrm{asym}}_{s,i}$ for the start of word $i$ and $\varepsilon^{\mathrm{asym}}_{e,i}$ for its end, is the distance from the predicted boundary to that interval (Appendix~\ref{app:metrics} gives the exact definition). A prediction inside the pause costs zero, and one that clips word $i$ or swallows part of its neighbor pays exactly the duration of speech clipped or swallowed. We report the mean asymmetric error over all boundaries.

\textbf{Word-level error rates.}
\label{sec:distribution}
Because a few catastrophic failures can skew the mean boundary error, studies also report the percentage of boundaries whose symmetric error falls within a tolerance, such as 10, 25, 50, or 100\,ms \citep{rousso2024tradition}. Each start and end boundary is thresholded independently, so a word with one bad edge still counts as half correct. An edit, however, fails if either cut clips speech or leaves an audible fragment (Figure~\ref{fig:eval-metrics}, middle). We therefore score each word by its worse edge, $\varepsilon_i=\max(\varepsilon^{\mathrm{asym}}_{s,i},\varepsilon^{\mathrm{asym}}_{e,i})$, and report the percentage of words with $\varepsilon_i$ above 25, 50, and 300\,ms.\footnote{We omit the 10\,ms tolerance because inter-annotator agreement on \bench{} is very low at that scale. Differences there reflect annotation noise more than aligner quality (Appendix~\ref{app:annotation}).} The 300\,ms rate captures catastrophic failures that may land in the wrong word.

\textbf{Scoring every word.}
\label{sec:coverage}
Both metrics above are often computed only over the words an aligner places; \citet{rousso2024tradition}, for example, compare aligners only on words that WhisperX and MMS recognized correctly. Aligners, however, differ in which words they place. MFA and MMS-FA place every transcript word, whereas Nyra and WhisperX can mark words not found depending on their tokenization (tokens Nyra cannot phonemize, or characters outside WhisperX's alphabet). Gentle reject words it cannot locate in the audio. Scoring only placed words therefore evaluates each aligner on its own set of words, which makes comparisons across aligners difficult, and lets a system that rejects difficult words report a deceptively low error on the rest (Figure~\ref{fig:eval-metrics}, bottom). We therefore score \emph{every} transcript word. Words detected by an aligner keep their spans, and each word marked $\nullint$ receives a fallback interval by dividing the gap between its nearest placed neighbors, or the audio boundaries, uniformly.\footnote{More elaborate fallbacks are possible, such as dividing the gap in proportion to each word's phoneme or character count rather than uniformly. We leave these to future work; the fallback only needs to be fixed and identical across systems for the comparison to be fair.} Every system is then scored on the same words: overconfident systems pay for bad boundaries, while selective systems pay for the fallback assigned to rejected words. Appendix~\ref{app:metrics} also reports the symmetric MAE for comparison with prior work.

\textbf{Evaluating under transcription errors.}
\label{sec:transcript-errors}
The metrics above are conventionally computed with hand-corrected transcripts that match the audio. To test robustness to realistic transcription errors, we also obtain transcripts from \scribe{} and \whisper{} and compare them with the ground-truth transcripts. Table~\ref{tab:asr-transcript-stats} summarizes the resulting substitution, deletion, and insertion rates, along with word error rate (WER) and character error rate (CER). For scoring aligners, we match each ASR word sequence to the ground-truth sequence by minimum edit distance. A word is \emph{matched} when the two transcripts agree on it. Boundary metrics use matched words only: a detected span is scored as usual, while a matched word marked not found receives the uniform interpolated boundary from Section~\ref{sec:coverage}. Insertions and substitutions have no valid reference interval, and deletions have no input token to place, so they are excluded from boundary metrics. Detection precision, recall, and F1 account for every input word supplied to the aligner: matched words marked found are true positives, matched words marked not found are false negatives, and inserted or substituted words marked found are false positives. Inserted or substituted words marked not found are true negatives. Appendix~\ref{app:transcript-errors} gives the full protocol. \looseness=-1

\section{The \bench{} Benchmark}
\label{sec:benchmark}

\textbf{Why a new benchmark?}
Existing benchmarks underrepresent long-form, accented, noisy, multi-speaker, and conversational speech (Appendix~\ref{app:corpora}). Table~\ref{tab:corpus-stats} compares them with \bench{}, which pools diverse recordings from public corpora. TIMIT contains 630 speakers reading ten sentences each in a noise-isolated booth with a headset microphone \citep{garofolo1993timit}. Buckeye contains 40 native speakers from central Ohio, interviewed in a quiet room with a head-mounted microphone \citep{pitt2007buckeye}.\footnote{Buckeye is distributed as roughly ten-minute sessions that include interviewer speech. We use its annotations to remove interviewer, overlapping, and third-speaker regions. We then retain single-speaker clips of at least 1\,s and three words (Appendix~\ref{app:reference-preparation}). All Buckeye statistics and results use this processed subset.} Both datasets contain only American English speakers. Their evaluation recordings are single-speaker and were collected under one condition. \bench{} instead covers varied recording conditions, multiple accents, long-form audio, and both single- and multi-speaker recordings. Multi-speaker clips include reference alignments for both speakers.

Text diversity matters because the aligner must connect written tokens to spoken words. We mark categories in which the two forms often differ: currency, numbers and years, times, acronyms, abbreviations, units, URLs, email addresses, symbols, and named entities. Table~\ref{tab:corpus-stats} reports the fraction of such \textit{difficult words} in each corpus. A word that belongs to several categories is counted once.

\begin{table}[t]
\centering
\begin{minipage}[t]{0.58\textwidth}
\vspace{0pt}
\centering
\scriptsize
\renewcommand{\arraystretch}{0.84}
\setlength{\tabcolsep}{2pt}
\caption{Evaluation-corpus statistics.}
\label{tab:corpus-stats}
\begin{tabular}{lrrr}
\toprule
 & \bench{} & TIMIT & Buckeye \\
\midrule
Speech style & mixed & read & conversational \\
Recording & mixed & booth/headset & quiet/headset \\
% Generated by eval/scripts/benchmark_stats.py; do not edit by hand.
Unique speakers & 73 & 630 & 40 \\
Non-American accent (\%) & 21.9 & 0.0 & 0.0 \\
Clips with $>1$ speaker (\%) & 27.4 & 0.0 & 0.0 \\
Difficult words (\%) & 9.6 & 2.4 & 5.9 \\
\midrule
Clips & 73 & 6,300 & 6,185 \\
Hours & 0.83 & 5.38 & 22.31 \\
Words & 8,072 & 54,387 & 244,408 \\
\bottomrule

\end{tabular}
\end{minipage}\hfill
\begin{minipage}[t]{0.40\textwidth}
\vspace{0pt}
\centering
\scriptsize
\renewcommand{\arraystretch}{0.84}
\setlength{\tabcolsep}{2pt}
\caption{ASR errors (\%); bold marks the lower WER/CER per dataset.}
\label{tab:asr-transcript-stats}
\begin{tabular}{llrrr}
\toprule
% Generated by aligner-evals/scripts/asr_metrics.py; do not edit by hand.
Dataset & ASR & Sub./Del./Ins. $\downarrow$ & WER $\downarrow$ & CER $\downarrow$ \\
\midrule
\bench{} & \scribe{} & 0.71 / 0.50 / 0.66 & \textbf{1.88} & \textbf{1.16} \\
 & \whisper{} & 3.14 / 8.76 / 2.76 & 14.65 & 9.78 \\
\midrule
TIMIT & \scribe{} & 1.12 / 0.35 / 0.23 & \textbf{1.71} & \textbf{0.38} \\
 & \whisper{} & 2.14 / 0.39 / 0.23 & 2.76 & 0.79 \\
\midrule
Buckeye & \scribe{} & 5.29 / 1.75 / 5.13 & \textbf{12.16} & \textbf{6.56} \\
 & \whisper{} & 4.69 / 8.02 / 3.38 & 16.09 & 10.16 \\
\bottomrule

\end{tabular}
\end{minipage}
\end{table}

\textbf{Design.}
To reflect an actual editing workload rather than public corpora's collection biases, we derive the target distribution from a private collection of real-world recordings processed by a widely deployed text-based editing system. This follows the use of reference distributions in ImageNetV2 \citep{recht2019imagenet} and real-world usage in Arena-Hard \citep{li2024crowdsourced} to guide benchmark construction. We select public clips to match aggregate audio, speaker, and transcript distributions while limiting repeated recordings and speakers. Selection was finalized before running any aligner on the candidate pool. Appendix~\ref{app:benchmark-details} describes the procedure.

\textbf{Statistics and annotation.}
\bench{} consists of 73 clips drawn from People's Speech \citep{galvez2021peoplesspeech}, VoxPopuli \citep{wang2021voxpopuli}, and AMI \citep{carletta2005ami}, all under licenses that permit redistribution (Appendix~\ref{app:licensing}). It has 8{,}072 words and 0.83 hours of audio with a median clip duration of 53.2\,s. 16 clips (21.9\%) have a non-American top accent, and 20 clips (27.4\%) contain more than one speaker. Difficult written forms, as defined above, make up 9.6\% of words. Three annotators independently place every word boundary. More details can be found in Appendix~\ref{app:annotation}. We will release the audio excerpts, verbatim transcripts, reference word boundaries, ASR transcripts, and evaluation code so that others can build on \bench{}.

\section{\sys{}}
\label{sec:model}
\sys{} separates global word localization from local boundary refinement. A joint audio-text transformer matches transcript words to coarse regions of the recording, while a convolutional head refines their start and end boundaries from local acoustic features. Operating the transformer on 16 ms audio patches keeps global attention tractable, after which convolutional upsampling produces boundary predictions on a 1 ms grid.

\textbf{Architecture.}
A strided 1D convolution maps a 16\,kHz waveform to $L$ audio tokens. It uses a 512-sample (32\,ms) window and a 256-sample (16\,ms) hop. In parallel, the frozen text encoder of the pre-trained Dia text-to-speech model supplies $C$ character token embeddings \citep{narilabs2025dia}. Dia was trained for text-to-speech, so its representations may capture how written words vary in pronunciation and prosody. We concatenate the $L+C$ tokens and pass them through a 12-layer transformer. Audio tokens receive RoPE positions $0,\dots,L-1$. Text token $k$ receives the fractional position $\frac{k}{C-1}\,L$, spreading the text uniformly over the same range and giving attention a soft diagonal prior between the two modalities.

We then split the transformer outputs by modality. Four convolutional stages independently upsample the audio representation for word starts and ends. They produce 1\,ms feature grids $y_s$ and $y_e$, each with $(2^4L=16L)$ frames. The text outputs are mean-pooled by word to obtain $W$ word vectors. Three MLPs map each vector to start and end classifiers $W_s,W_e$ and a presence probability $p$. The products $W_s y_s^\top$ and $W_e y_e^\top$ form $W\times16L$ boundary heatmaps. Each word, and each of its two boundaries, is localized independently. A transcript or pronunciation mismatch therefore need not shift neighboring words, and the presence head can reject words that are absent from the audio. Appendix~\ref{app:model-details} specifies the dimensions, heads, and decoding.

\begin{figure}[t]
\centering
\includegraphics[width=\textwidth]{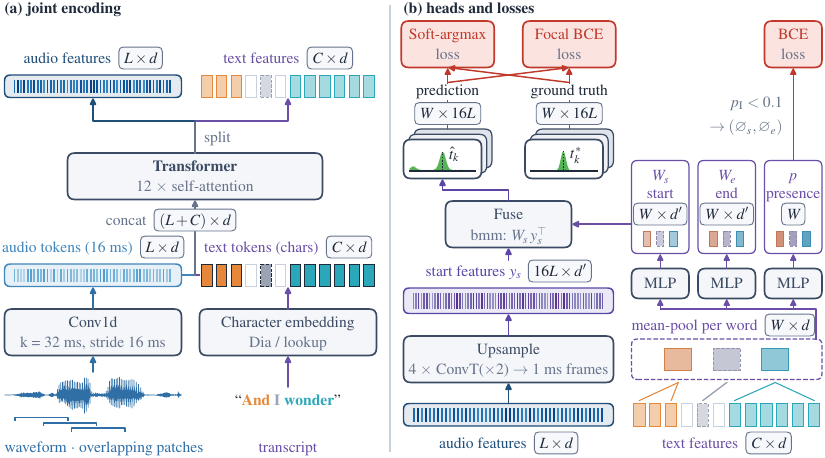}
\caption{\textbf{\sys{} architecture.} Audio patches and character tokens are jointly contextualized by a transformer; convolutional refinement then produces millisecond-resolution word-boundary heatmaps, while a presence head rejects absent words (Section~\ref{sec:model}).}
\label{fig:architecture}
\end{figure}

\textbf{Presence head.}
\label{sec:presence-head}
ASR transcripts can contain words that are absent from the audio. A globally constrained decoder must place every accepted token, so one mismatch can shift neighboring intervals. \sys{} avoids this propagation by localizing words independently, but it could still place an unsupported word on a similar-sounding frame. We therefore predict $p(\text{spoken}\mid\text{word},\text{audio})$. During training, we insert random transcript words at a 5\% rate and mask their boundary loss.\footnote{This rate roughly matches the insertion errors produced by Scribe on our internal data. Future work could also simulate deletions and substitutions.} A class-balanced BCE trains the presence head. At inference, $p<0.1$ produces $\nullint$ and invokes the fallback from Section~\ref{sec:coverage}.

\textbf{Training objective.}
Each word $k$ has a start heatmap with 1\,ms-frame logits $z_{k,t}$ and labeled frame $t^\ast_k$; the end heatmap is trained identically. A one-hot target is too sparse (one positive among ${\sim}3\times10^4$ frames in a 30\,s clip), and Gentle's labels jitter on a 10\,ms grid. We therefore use a Gaussian target $g_{k,t}=\exp\!\big(-(t-t^\ast_k)^2/2\sigma^2\big)$ with $\sigma=10$\,ms under focal BCE \citep{lin2017focal} extended to soft targets, which down-weights easy negatives:
\begin{equation}
\mathcal{L}_{\mathrm{focal}}=\frac{\sum_{k,t}\mathrm{FocalBCE}\big(p_{k,t},\,g_{k,t}\big)}{\sum_{k,t} g_{k,t}},\qquad p_{k,t}=\operatorname{sigmoid}(z_{k,t}).
\label{eq:focal}
\end{equation}
Once a predicted peak falls outside the Gaussian, the target near it is zero whether the miss is moderate or large, so this loss carries little distance information. We therefore add a loss on the soft-argmax $\hat t_k$, the expected boundary under the predicted distribution, with the displacement normalized by the clip length $T$:
\begin{equation}
\hat t_k=\sum_t \operatorname{softmax}_t(z_{k,\cdot})\,t,\qquad
\mathcal{L}_{\mathrm{argmax}}=\frac{1}{W}\sum_{k=1}^{W}\ell\Big(\frac{\hat t_k-t^\ast_k}{T}\Big).
\label{eq:argmax}
\end{equation}
Here $\ell$ is smooth-$L_1$, linear beyond $\beta T$ (300\,ms for a 30\,s clip), so catastrophic errors pull distant peaks toward the label without dominating the gradient. The presence head uses class-balanced BCE. Appendix~\ref{app:loss} defines $\mathrm{FocalBCE}$ and gives the coefficients.

\textbf{Data.}
We begin with a large corpus of transcribed but unaligned speech and use Gentle \citep{ochshorn2016gentle} to create pseudo-labeled word boundaries. We train \sys{} for 100k steps with a batch size of 128, corresponding to roughly 78,000 hours of audio. Gentle occasionally collapses a word onto one to three frames. We mask boundary supervision for words shorter than 30\,ms, except short words that plausibly occur in continuous speech; Appendix~\ref{app:pseudo-label-filtering} gives the rule.

\textbf{EMA correction.}
Manual review found errors in Gentle's labels, and we observed that, after first learning the consistent labels, the model begins to fit these errors and makes more catastrophic mistakes. We therefore correct labels online, using a rolling bank of EMA model snapshots as teachers: a Gentle label is kept when the teachers agree with it, replaced by their median when they agree with one another but not with Gentle, and masked otherwise. Appendix~\ref{app:training-curve} gives details.

\textbf{Augmentation.}
We also augment the waveform with reverberation, noise, filtering, clipping, and gain changes. These corruptions expose the model to a wider range of acoustic conditions. Interested readers can refer to Appendix~\ref{app:model-details} for more details.

\textbf{Long-form inference.}
\sys{} processes at most 30\,s per model call to keep joint audio-text attention tractable. For longer recordings, we use approximate ASR word timestamps to route transcript words into overlapping, speech-aware audio chunks. We use caller-supplied times when available and otherwise run \whisper{}. These times guide chunking only; \sys{} predicts every final boundary. Appendix~\ref{app:long-form} describes routing and stitching.

\section{Experiments}
\label{sec:experiments}

\subsection{Setup}
\label{sec:baselines}
\textbf{Data and baselines.}
We evaluate on \bench{}, TIMIT, and Buckeye. Each recording has three transcripts: the hand-corrected reference and the unedited outputs of \scribe{} and \whisper{}, scored using the protocol in Section~\ref{sec:transcript-errors}.
Our baselines cover two dominant families: lexicon-and-HMM aligners which includes Gentle \citep{ochshorn2016gentle}; MFA 3.1 with its public English models \citep{mcauliffe2017montreal,mcauliffe2026montreal}; and Nyra with its default beam ladder \citep{nyralabs2026forcedaligner}\footnote{Nyra's 159-hour training mix includes AMI, which supplies 11 of \bench{}'s 73 clips. Exact training overlap is unknown, so Nyra's scores may be optimistic on this subset.} and CTC aligners: MMS-FA through torchaudio's \texttt{MMS\_FA} pipeline \citep{pratap2024scaling}, and WhisperX's English wav2vec\,2.0 aligner, without its transcription step \citep{bain2022whisperx}.  We do not run speech-LLM aligners such as Qwen3-ForcedAligner \citep{qwen3asr2026} because it's output timestamp resolution is substantially coarser than the error scale of our aligners (Table~\ref{tab:development-results}). None of the evaluation recordings, nor any recording from their source corpora (People's Speech, VoxPopuli, AMI, TIMIT, Buckeye), is part of \sys{} training data. \looseness=-1

\textbf{Protocol.}
\label{sec:textnorm}
Baselines run on whole clips. \sys{} is trained on chunks shorter than 30\,s, so on longer recordings, including most of \bench{}, it must chunk the audio and stitch the per-chunk outputs as described in Section~\ref{sec:model}. We use Whisper to obtain coarse timestamps for this process. Its errors, including errors near chunk boundaries, remain part of the reported results. All systems receive the same transcripts, normalized with NeMo's WFST text normalizer \citep{zhang2021nemo}. Before alignment, each written token is expanded to spoken form. We then map it back to the interval from the first expanded word's start to the last expanded word's end.\footnote{The token ``2001'' expands to ``two thousand one''; the start of ``two'' and the end of ``one'' are reported for ``2001.'' Appendix~\ref{app:textnorm} gives details.}

\subsection{Main results}
\label{sec:main-results}
Table~\ref{tab:development-results} reports every system on the three corpora; each cell reads left to right from the reference-consistent setting to the real-ASR setting. For comparison with prior work, Table~\ref{tab:symmetric-errors} in Appendix~\ref{app:metrics} reports the conventional symmetric error metric: the mean absolute boundary error averaged over all start and end boundaries, without the silence tolerance.

\begin{table}[ht]
\centering
\small
\caption{Main results. Each cell is \emph{hand-corrected / \scribe{} / \whisper{}} for the same audio. $\downarrow$ lower is better, $\uparrow$ higher is better. We score exact ASR/reference word matches only, with the uniform fallback for unplaced words (Section~\ref{sec:metrics}). Within each dataset and at each column position, bold marks the best system and every system statistically tied with it: each point estimate lies within the other's 95\% word-level confidence interval. F1 has no interval, so only its best value is bold.}
\label{tab:development-results}
\resizebox{\textwidth}{!}{%
\begin{tabular}{llrrrrr}
\toprule
Dataset & System & \% ${>}300$\,ms $\downarrow$ & \% ${>}50$\,ms $\downarrow$ & \% ${>}25$\,ms $\downarrow$ & Mean asym.\ (ms) $\downarrow$ & F1 (\%) $\uparrow$ \\
\midrule
\bench{} & \sys{} & \textbf{0.38} / \textbf{0.43} / \textbf{1.71} & \textbf{7.27} / \textbf{7.18} / \textbf{7.74} & \textbf{29.82} / \textbf{29.87} / \textbf{29.77} & \textbf{17.16} / \textbf{20.43} / \textbf{32.82} & 99.82 / 99.28 / 96.83 \\
 & Gentle & 2.11 / 2.04 / 3.67 & 12.61 / 12.53 / 13.48 & 35.20 / 35.13 / 35.64 & 31.43 / 33.68 / 51.46 & 98.90 / 98.26 / \textbf{97.31} \\
 & MFA & 4.15 / 4.89 / 10.82 & 34.16 / 34.52 / 38.56 & 64.92 / 65.18 / 67.18 & 73.00 / 79.80 / 374.26 & \textbf{100.00} / 98.20 / 95.20 \\
 & Nyra & 1.28 / 1.26 / 2.57 & 29.99 / 29.85 / 28.88 & 68.84 / 68.77 / 68.02 & 33.18 / 33.42 / 56.90 & \textbf{100.00} / 99.31 / 96.81 \\
 & MMS-FA & 2.49 / 2.43 / 3.84 & 46.69 / 46.85 / 45.73 & 80.74 / 81.14 / 80.73 & 48.91 / 50.44 / 73.89 & 98.96 / 99.31 / 96.82 \\
 & WhisperX & 1.41 / 1.66 / 3.57 & 53.25 / 53.16 / 52.44 & 84.75 / 84.69 / 84.43 & 49.04 / 49.84 / 90.38 & 99.78 / \textbf{99.33} / 96.81 \\
\midrule
TIMIT & \sys{} & \textbf{0.02} / \textbf{0.02} / \textbf{0.02} & 8.60 / 8.49 / 8.62 & 51.06 / 50.96 / 50.91 & 18.37 / 18.32 / 18.29 & 99.99 / 99.29 / 98.76 \\
 & Gentle & 0.34 / 0.31 / 0.31 & 13.70 / 13.58 / 13.59 & 55.71 / 55.54 / 55.49 & 22.45 / 22.23 / 22.24 & 99.80 / 99.16 / 98.64 \\
 & MFA & 0.16 / 0.18 / 0.18 & 17.08 / 17.06 / 17.17 & 39.06 / 39.11 / 39.14 & 18.21 / 18.23 / 18.26 & \textbf{100.00} / \textbf{99.29} / 98.76 \\
 & Nyra & \textbf{0.01} / \textbf{0.01} / \textbf{0.01} & \textbf{8.01} / \textbf{7.99} / \textbf{8.05} & \textbf{28.84} / \textbf{28.81} / \textbf{28.61} & \textbf{13.67} / \textbf{13.67} / \textbf{13.62} & \textbf{100.00} / \textbf{99.29} / 98.76 \\
 & MMS-FA & \textbf{0.02} / \textbf{0.02} / \textbf{0.02} & 47.96 / 47.43 / 45.91 & 81.81 / 81.44 / 80.19 & 36.26 / 35.55 / 33.79 & \textbf{100.00} / \textbf{99.29} / \textbf{98.76} \\
 & WhisperX & 0.16 / 0.14 / 0.12 & 65.16 / 64.93 / 64.66 & 89.20 / 89.12 / 89.04 & 45.34 / 44.90 / 43.85 & \textbf{100.00} / \textbf{99.29} / 98.76 \\
\midrule
Buckeye & \sys{} & \textbf{0.40} / \textbf{0.26} / \textbf{1.27} & \textbf{10.52} / \textbf{10.04} / \textbf{10.90} & 51.34 / 51.25 / 52.50 & 22.70 / 21.10 / \textbf{30.64} & 99.54 / 94.55 / 95.47 \\
 & Gentle & 1.16 / 1.02 / 2.48 & 14.02 / 12.78 / 14.24 & 52.75 / 51.99 / 53.33 & 40.42 / 34.93 / 46.66 & 99.41 / \textbf{94.75} / \textbf{95.55} \\
 & MFA & 1.98 / 1.65 / 5.78 & 21.93 / 21.26 / 26.00 & 44.62 / 44.00 / 47.55 & 34.06 / 32.99 / 103.15 & \textbf{99.99} / 94.55 / 95.33 \\
 & Nyra & 0.59 / 0.45 / 2.06 & 12.45 / 12.23 / 13.71 & \textbf{37.55} / \textbf{37.26} / \textbf{38.28} & \textbf{21.62} / \textbf{19.93} / 37.14 & \textbf{99.99} / 94.55 / 95.44 \\
 & MMS-FA & 1.35 / 1.18 / 2.88 & 50.26 / 50.66 / 49.80 & 84.54 / 84.98 / 84.65 & 47.20 / 45.55 / 58.31 & 99.92 / 94.52 / 95.44 \\
 & WhisperX & 1.41 / 1.17 / 2.71 & 64.76 / 64.91 / 65.29 & 91.43 / 91.76 / 91.90 & 54.33 / 52.06 / 71.36 & \textbf{99.99} / 94.52 / 95.44 \\
\bottomrule

\end{tabular}
}
\end{table}

\textbf{Results with hand-corrected transcripts.}
\sys{} performs best on \bench{}, with a catastrophic-error rate approximately $3-11\times$ lower than the baselines on its long-form, noisy, accented, and multi-speaker recordings. Nyra instead leads on clean read TIMIT speech, while Buckeye lies between these regimes: \sys{} is better in the tail, whereas Nyra is stronger at finer tolerances and in mean error. This suggests a consistent tradeoff: HMM-based aligners with strong acoustic front-ends remain highly precise on narrow, carefully annotated speech, while \sys{} is more robust on varied in-the-wild recordings. CTC aligners are consistently coarser across datasets. Table~\ref{tab:subgroup-results} further breaks down \bench{} by accent, speaker count, and word difficulty.

\textbf{On ASR transcripts.}
With \scribe{}, whose error rates are relatively low, alignment performance changes little; small apparent improvements can occur because boundary metrics are computed only on exact ASR-reference matches, which removes some difficult words from evaluation. In contrast, the noisier \whisper{} transcripts primarily increase catastrophic alignment errors while leaving the 25\,ms error rate nearly unchanged. This indicates that transcription errors affect the tail of the alignment distribution more than fine-grained boundary precision. Under this setting, \sys{} degrades least and retains the lowest catastrophic error rates and mean error on both \bench{} and Buckeye, whereas globally constrained HMM systems, especially MFA, suffer much larger tail failures. Overall, the advantage of \sys{} is largest when both the transcript and recording conditions depart from the clean benchmark setting.

\textbf{Detection F1.}
Detection F1 and boundary accuracy capture different failure modes. High detection performance does not imply accurate timing: for example, MFA marks nearly every word found on \bench{} despite much larger boundary errors. Conversely, under \whisper{}, \sys{} has slightly lower F1 than Gentle while remaining substantially more accurate in timing. Detection alone is therefore insufficient for alignment quality assessment and should be reported alongside boundary accuracy.

\subsection{Ablations}
\label{sec:ablations}
We ablate a few important components of \sys{} in Table~\ref{tab:model-ablations} and report results on \bench{} with the hand-corrected transcript. We use the hand-corrected transcript for the main ablations to isolate changes in boundary localization from changes caused by ASR errors; Appendix~\ref{app:model-ablations-asr} reports the same ablations under \scribe{} and \whisper{} transcripts. The baseline uses audio augmentation, EMA-snapshot label correction, a 12-layer 1024-wide transformer with a 4096-wide FFN, Dia text features, a 512-sample patch window with 256-sample hop, and batch size 128.

\begin{table}[ht]
\centering
\small
\caption{Model ablations on \bench{} (hand-corrected transcript). Parameter counts cover only the trainable alignment model and exclude the frozen 252M-parameter Dia encoder. Small $\pm$ values are 95\% word-level confidence half-widths.}
\label{tab:model-ablations}
\resizebox{\textwidth}{!}{%
\begin{tabular}{llrrrr}
\toprule
% Generated by eval/scripts/paper_tables.py; do not edit by hand.
% run 136 <- 136-baseline-fixed_warmup-fixed_lambda_100000 (8072 words)
% -- Model hyperparameters --
% run 123 <- 123-baseline_121-small_model_100000 (8072 words)
% run 126 <- 126-baseline_121-bigboi_100000 (8072 words)
% run 130 <- 130-baseline_121-huge-crosstalk_0p5-aug_0p5-200k_steps_100000 (8072 words)
% run 127 <- 127-baseline_121-patch_512_1024_100000 (8072 words)
% run 125 <- 125-baseline_121-char_tokenizer_100000 (8072 words)
% run 128 <- 128-baseline_121-bs64_100000 (8072 words)
% -- Augmentation --
% run 122 <- 122-baseline_121-augmentation_off_100000 (8072 words)
% run 129 <- 129-baseline_121-crosstalk_100000 (8072 words)
% -- Label correction --
% run 133 <- 133-baseline_121-no_ema-bs8-gradacc2_100000 (8072 words)
% -- Loss and output resolution --
% run 134 <- 134-baseline_121-no_argmax_loss_100000 (8072 words)
% run 135 <- 135-baseline_121-output_ms_16_100000 (8072 words)
Ablation & Change from the baseline & Mean asym.\ (ms) $\downarrow$ & \% ${>}300$\,ms $\downarrow$ & \% ${>}50$\,ms $\downarrow$ & \% ${>}25$\,ms $\downarrow$ \\
\midrule
-- & baseline (${\sim}213$M params) & 17.16\ci{1.92} & 0.38\ci{0.13} & 7.3\ci{0.6} & 29.8\ci{1.0} \\
\midrule
1 & width = 512, FFN = 2048 (${\sim}60$M params) & 17.75\ci{2.49} & 0.35\ci{0.13} & 7.1\ci{0.6} & 29.6\ci{1.0} \\
2 & 24 layers (${\sim}415$M params) & 17.01\ci{1.91} & 0.33\ci{0.13} & 7.4\ci{0.6} & 30.1\ci{1.0} \\
3 & width = 2048, FFN = 8192 (${\sim}821$M params) & 22.23\ci{4.62} & 0.33\ci{0.13} & 7.3\ci{0.6} & 30.0\ci{1.0} \\
4 & window = 1024, hop = 512 samples & 17.77\ci{2.48} & 0.32\ci{0.12} & 7.3\ci{0.6} & 29.9\ci{1.0} \\
5 & learned character embeddings (instead of frozen Dia) & 17.67\ci{1.94} & 0.40\ci{0.14} & 7.8\ci{0.6} & 30.0\ci{1.0} \\
6 & batch size = 64 & 17.21\ci{1.92} & 0.37\ci{0.13} & 7.6\ci{0.6} & 29.8\ci{1.0} \\
\midrule
7 & no audio augmentation & 18.05\ci{2.49} & 0.32\ci{0.12} & 7.7\ci{0.6} & 30.5\ci{1.0} \\
8 & crosstalk augmentation & 17.98\ci{2.54} & 0.35\ci{0.13} & 7.5\ci{0.6} & 30.0\ci{1.0} \\
\midrule
9 & EMA-snapshot correction off & 19.46\ci{2.59} & 0.68\ci{0.18} & 8.3\ci{0.6} & 30.9\ci{1.0} \\
\midrule
10 & no soft-argmax loss & 19.83\ci{3.45} & 0.38\ci{0.13} & 7.7\ci{0.6} & 30.1\ci{1.0} \\
11 & 16\,ms output grid (no convolutional upsampling) & 21.62\ci{4.41} & 0.41\ci{0.14} & 7.8\ci{0.6} & 31.7\ci{1.0} \\
\bottomrule

\end{tabular}
}
\end{table}

Model scale and most architectural choices have little effect. Models from 60M to 415M parameters perform similarly (ablations 1 - 2), while the 821M model is worse (ablation 3); patch size, text representation, batch size, and audio augmentation (ablations 4 - 8) also produce only modest changes. Thus, performance does not improve monotonically with model capacity.

The components that matter instead target different failure modes. EMA-snapshot label correction (ablation 9) primarily suppresses catastrophic errors, supporting its role in preventing overfitting to noisy pseudo-labels. The soft-argmax loss (ablation 10) reduces large boundary displacements, while convolutional upsampling (ablation 11) yields the largest improvement at fine temporal tolerances. Together, these results suggest that \sys{} benefits more from controlling label noise and explicitly modeling boundary precision than from increasing model scale. Appendix~\ref{app:conditions} separately ablates the evaluation protocol.

\section{Conclusion}
Existing forced-alignment evaluations miss important deployment conditions, particularly diverse long-form audio and imperfect transcripts. We address this with an evaluation framework that captures editing-relevant failures, a new diverse benchmark  and \sys{}, a forced aligner that predicts word boundaries independently at fine temporal resolution without a fixed lexicon path. Across clean and ASR-generated transcripts, the results show that robustness to large alignment failures is not captured well by conventional clean benchmarks alone. \sys{} is strongest on the varied, in-the-wild conditions targeted by \bench{}, while ablations indicate that label correction and fine-resolution boundary refinement matter more than model scale.

\sys{} still has two limitations. First, training requires pseudo-aligned speech, which may be difficult to obtain in languages without a suitable bootstrap aligner; EMA label correction reduces sensitivity to label noise but does not remove this dependency. Second, long-form inference relies on coarse timestamps to route transcript segments into model-sized chunks, so it is not fully self-contained. These timestamps can often be supplied by the ASR stage already present in editing and data-preparation pipelines. Future work could focus on reducing the need for strong bootstrap aligners and integrating long-form routing directly into the alignment model.

% ICLR checklist items. arXiv does not require them; licensing and annotation
% details remain in the appendix for both builds.
\ifdefined\arxivpreprint\else
\clearpage % added this so we know where 9 page limit ends

\subsubsection*{AI Use Statement}
In this work, we used generative AI tools to implement methods, including portions of the model, training, and evaluation code. We have not used generative AI tools to generate synthetic data sets, develop theoretical models or conceptual frameworks, propose or refine hypotheses, design or provide feedback on research methodology or experiments, clean or reformat datasets, or interpret results. Formulating or proving mathematical claims, writing proofs, translation, qualitative or thematic data analysis, and formulating survey or interview questions are not applicable to this work. Additionally, we used generative AI tools to create and edit software code, to create and modify scientific figures, to create research artifacts, to draft parts of the paper and edit it for readability, and to search for information, identify relevant literature, and summarize existing literature. We have reviewed all AI-assisted work: the authors reviewed and tested AI-assisted code, verified all discovered sources and citations against the original publications, and checked all AI-assisted text and figures against the underlying research and results. We take responsibility for the final content of this work, including text, claims, and artifacts produced with the aid of generative AI.

\subsubsection*{Ethics Statement}
All released \bench{} audio comes from public corpora whose licenses permit redistribution: CC BY 4.0 for AMI, CC0 1.0 for VoxPopuli, and item-level CC BY 2.0 - 4.0 for the selected People's Speech recordings. The audio remains under those licenses, while our annotations and code are released under CC BY 4.0. Appendix~\ref{app:licensing} lists per-source attribution. The released manifest records each clip's creator, license, source URL, and modifications. We use the private recordings that define the target distribution only to estimate aggregate marginals; we do not release them. Appendix~\ref{app:annotation} describes annotator sourcing, compensation, agreement, adjudication, and the review of public clips for personally identifiable information.

\subsubsection*{Reproducibility Statement}
Upon acceptance, we will release \bench{} and the evaluation code. Sections~\ref{sec:metrics} and \ref{sec:experiments} define the metrics and experimental protocol. Appendix~\ref{app:benchmark-details} describes benchmark construction, Appendix~\ref{app:model-details} gives model and training details, and Appendix~\ref{app:reference-preparation} documents reference-corpus preparation.
\fi

\ificlrfinal
\subsubsection*{Acknowledgments}
We thank Heba Asmar and Lorenzo Simionato for helpful discussions and feedback.
\fi

\bibliography{references}
\bibliographystyle{iclr2027_conference}

\newpage

\appendix
\section*{Appendix}

\section{Additional Related Work}
\label{app:related}
\label{app:corpora}

\textbf{HMM-based aligners.}
The classical family also includes P2FA \citep{yuan2008speaker}, FAVE \citep{rosenfelder2011fave}, Prosodylab-Aligner \citep{gorman2011prosodylab}, and the web-hosted WebMAUS \citep{kisler2012signal}. MFA and Gentle are their current Kaldi-based successors. Gentle's fixed English model and vocabulary cannot be adapted and boundaries remain on the recognizer's frame grid. Gentle occasionally collapses a word onto one to three frames, which motivates the pseudo-label filtering in Section~\ref{sec:model}. Neither Gentle nor MFA learns from a word-boundary objective; their precision follows from the acoustic model's frame rate, while their robustness depends on the lexicon and acoustic training data.

The Nyra forced aligner is the closest recent successor of this family \citep{nyralabs2026forcedaligner}. It retains three-state left-to-right triphone HMMs trained only from its own alignments, with no external boundary labels. The changes are the front-end and the inventory: WavLM-Large layer-22 states, interpolated to 10\,ms and reduced first by PCA and then by a noise-paired LDA, plus filler and vocal-event units supervised from verbatim tags. The authors report 20.6\,ms mean word-boundary error on Buckeye against MFA \texttt{us\_arpa} at 25.3\,ms under a widened beam, with $2.7\times$ fewer errors beyond 200\,ms. Our protocol differs (asymmetric per-word max, all-word fallback, processed Buckeye turns rather than their segmentation, and MFA's public Global-English bundle), so the absolute numbers are not a replication, but Table~\ref{tab:development-results} agrees on the qualitative ranking: Nyra is the strongest HMM system on TIMIT and Buckeye. On \bench{} it remains far coarser than \sys{} (30\% of words above 50\,ms vs.\ 7\%), which is consistent with a global phone path that was not trained on long, multi-speaker, in-the-wild media. Running it on \sys{}'s chunks lowers this to 20\% (Table~\ref{tab:conditions}), so about half of the gap to \sys{} remains once clip length is controlled for.

\textbf{CTC-based neural aligners.}
A second family runs a CTC-trained ASR model to obtain framewise token posteriors, then finds a transcript-constrained path through them: ctc-segmentation \citep{kurzinger2020ctc}, NeMo NFA \citep{rastorgueva2023nemo}, and Charsiu \citep{zhu2022phone}. These systems differ in backbone and vocabulary (characters, phones, or sub-words) but share the decoding mechanism, and their boundary precision is bounded by the same frame rate and peaky CTC posteriors. We therefore evaluate the family through two representatives with public checkpoints: torchaudio's MMS-FA pipeline \citep{pratap2024scaling}, a character-level model trained on 1,000+ languages, and WhisperX's English letter-level wav2vec\,2.0 aligner \citep{bain2022whisperx}.

\textbf{Speech-LLM aligners.}
Most recently, LLM-ForcedAligner \citep{mu2026llm} and Qwen3-ForcedAligner \citep{qwen3asr2026} recast alignment as slot filling: timestamp slots are inserted into the transcript and a speech LLM predicts a discrete time index per slot, non-autoregressively. These models bring multilinguality and long-form robustness, and report large reductions in accumulated shift versus prior neural methods. However, the output is quantized to the encoder hop (e.g. 80\,ms for Qwen) over a fixed maximum duration - coarser than the \emph{mean} error of classical aligners, and is hence appropriate for tasks like subtitling but not accurate enough for editing.

\textbf{Attention-supervised Whisper timing.}
CrisperWhisper~2.0 \citep{wagner2026transcription} does not add an alignment module. After making the output policy verbatim, so that every spoken event has a token and word timing is well-defined, it finds the ten cross-attention heads whose untrained patterns already correlate with MFA word intervals and trains only those. Supervision is a cosine-similarity loss between the heads' average attention and a binary occupancy vector over encoder frames; supervising the average rather than each head lets them specialize on different parts of the word. Decoding pools the word's rows, sharpens the peak, and runs a monotonic Viterbi path that alternates word states with a mel-energy blank (shaped by \texttt{blank\_gamma} and \texttt{blank\_penalty} so quiet interiors of words are not eaten). The encoder hop is 20\,ms. The public \texttt{forced\_align} entry point still transcribes with this decoder, then copies times onto the reference via \texttt{SequenceMatcher} and interpolates unmatched words; it does not teacher-force the supplied text and is hence not included in our evaluation.

\textbf{Additional benchmark corpora.}
Beyond TIMIT and Buckeye, several corpora provide detailed manual timing, including MOCHA-TIMIT \citep{wrench2000mocha}, mngu0 \citep{richmond2011announcing}, USC-TIMIT \citep{narayanan2014real}, and Wisconsin XRMB \citep{westbury1994x}. These resources are small, usually contain a handful of speakers reading scripted prompts, and were collected primarily to study articulation through EMA, palatography, real-time MRI, or X-ray microbeam measurements. They are valuable for articulatory and phone-level analysis, but do not exercise the conversational, long-form, noisy, or multi-speaker conditions targeted by our word-alignment benchmark, so we do not include them as additional evaluation sets.

\section{Detailed Evaluation Metrics}
\label{app:metrics}

\textbf{Silence-aware boundary error.}
For prediction $(\hat{s}_i,\hat{e}_i)$ and reference $(s_i,e_i)$, symmetric error is the absolute displacement of each edge, $|\hat{s}_i-s_i|$ and $|\hat{e}_i-e_i|$. Asymmetric error penalizes only the part of that displacement that lies inside speech (Section~\ref{sec:boundary}). For the start of word $i$, whose predecessor ends at $e_{i-1}$, and for its end, whose successor starts at $s_{i+1}$,
\begin{align*}
\varepsilon^{\mathrm{asym}}_{s,i} &= \max(0,\,\hat{s}_i-s_i) + \max(0,\,\min(e_{i-1},s_i)-\hat{s}_i),\\
\varepsilon^{\mathrm{asym}}_{e,i} &= \max(0,\,e_i-\hat{e}_i) + \max(0,\,\hat{e}_i-\max(s_{i+1},e_i)).
\end{align*}
The first term of each expression is the part of the word that the boundary clips; the second is the part of the neighboring word, outside word $i$ itself, that it swallows. Equivalently, each error is the distance from the predicted boundary to a free interval, $[\min(e_{i-1},s_i),\,s_i]$ for a start and $[e_i,\,\max(s_{i+1},e_i)]$ for an end. Any position inside the adjacent pause costs zero, and a boundary that crosses into the neighbor is penalized the overlap rather than its full displacement. Reference words can overlap. For example, 52 of the 7{,}999 \bench{} junctions overlap by more than 5\,ms through crosstalk (Appendix~\ref{app:annotation}), and 870 of the 48{,}087 junctions in TIMIT's word tier overlap. The part of the neighbor that overlaps word $i$ cannot be excluded without clipping word $i$, so it is not penalized. The free interval then collapses to the reference boundary, the error reduces to the symmetric displacement, and an exact prediction scores zero. Without the clamp, every prediction at an overlapping junction, including the reference itself, would pay at least the overlap width. At most one term is ever non-zero. For the first word we set $e_{0}=-\infty$ and for the last word $s_{N+1}=+\infty$, so the lead-in before the first word and the tail after the last are free; an early first start or a late last end therefore costs nothing, while a late first start or an early last end still pays for the speech it removes. The reported mean asymmetric error pools all $2N$ start and end boundaries.

\textbf{Word-level error rates.}
Each word is scored by its worse edge,
$\varepsilon_i=\max(\varepsilon^{\mathrm{asym}}_{s,i},\varepsilon^{\mathrm{asym}}_{e,i})$,
because editing imposes a conjunction: both cuts around a word must be acceptable. For example, a word with start and end errors of $(0,40)$\,ms and one with $(20,20)$\,ms have the same 20\,ms mean, but at a 25\,ms tolerance only the first clips audible speech. Thresholding each edge independently counts the first word as half correct; thresholding its worse edge fails it, as in Figure~\ref{fig:eval-metrics} (middle). We report the percentage of words with $\varepsilon_i$ above 25, 50, and 300\,ms.

\textbf{Scoring every word.}
Every transcript word is scored. Placed words keep their predicted intervals. Each maximal run of consecutive words marked $\nullint$ receives the gap between its nearest placed neighbors, divided evenly among its words. A leading run starts at zero, a trailing run ends at the audio duration, and a transcript with no placed words divides the whole clip. The fallback does not alter placed intervals: they are neither made monotonic nor clamped to the audio duration.

\textbf{Symmetric results.}
Table~\ref{tab:symmetric-errors} reports the conventional symmetric mean for every cell of Table~\ref{tab:development-results}, pooled over all start and end boundaries of every scored word. Because it preserves absolute start and end displacement, it is useful when predicted boundaries are used to analyze word durations. With hand-corrected transcripts, \sys{} is the best system on \bench{}, while Nyra is best on TIMIT and Buckeye. That ranking persists with both ASR transcripts on TIMIT; on Buckeye under \whisper{}, \sys{} has the lower symmetric mean, hinting towards the robustness of \sys{}.

\begin{table}[ht]
\centering
\small
\caption{Mean symmetric boundary error (ms)~$\downarrow$ for every evaluation dataset and aligner. Each cell is \emph{hand-corrected / \scribe{} / \whisper{}}, matching Table~\ref{tab:development-results}; bold marks the best system at each transcript position within a dataset.}
\label{tab:symmetric-errors}
\resizebox{\textwidth}{!}{%
\begin{tabular}{lrrrrrr}
\toprule
Dataset & \sys{} & Gentle & MFA & Nyra & MMS-FA & WhisperX \\
\midrule
\bench{} & \textbf{21.51} / \textbf{25.46} / \textbf{38.11} & 37.59 / 39.74 / 57.44 & 80.46 / 87.38 / 383.63 & 37.83 / 38.31 / 61.88 & 80.71 / 81.35 / 99.57 & 54.20 / 53.55 / 94.85 \\
TIMIT & 27.65 / 27.60 / 27.66 & 28.82 / 28.63 / 28.69 & 23.23 / 23.26 / 23.33 & \textbf{23.06} / \textbf{23.10} / \textbf{23.13} & 37.17 / 41.06 / 48.05 & 46.16 / 45.80 / 44.87 \\
Buckeye & 30.51 / 27.62 / \textbf{37.73} & 48.24 / 41.57 / 53.81 & 42.27 / 38.65 / 112.51 & \textbf{27.66} / \textbf{25.08} / 43.32 & 53.22 / 70.87 / 80.32 & 59.80 / 55.44 / 76.08 \\
\bottomrule

\end{tabular}
}
\end{table}

\section{Transcript-Error Taxonomy and ASR Protocol Details}
\label{app:transcript-errors}

Transcript errors come in three flavors, each stressing the aligner differently:
\begin{itemize}[leftmargin=*, itemsep=1pt]
  \item \textbf{Insertions} (word in transcript, not in audio): can confuse the aligner into assigning audio to a phantom word, corrupting its neighbors. ASR hallucinations are an example.
  \item \textbf{Deletions} (word in audio, not in transcript): the acoustic evidence for the deleted word must be absorbed somewhere; a brittle aligner latches onto it and marks a real neighboring word missing. Some ASR models especially miss out on faint speech or disfluencies.
  \item \textbf{Substitutions} (wrong word in transcript): these include spelling variants, \emph{can't}/\emph{cannot}, and misspelt proper nouns. The wrong token has no valid reference boundary, but a robust presence detector should avoid claiming that it was found.
\end{itemize}

We transcribe each audio independently with \whisper{} and \scribe{} and retain each system's output without human correction. Each aligner is then run three times: once with the hand-corrected transcript and once with each ASR transcript.

For scoring, ASR and ground-truth word sequences are lowercased, stripped of punctuation, text-normalized and aligned by minimum edit distance; among the minimum-edit alignments we take the one with the most exact matches. Detection metrics are computed over every ASR word the aligner received. An exact matched pair is a true positive when the aligner marks it found and a false negative otherwise. A substituted or inserted ASR word is a false positive when marked found and a true negative otherwise. Deleted ground-truth words are ignored because the aligner never received them. Precision is therefore $\mathrm{TP}/(\mathrm{TP}+\mathrm{FP})$ with found substitutions and insertions in the denominator, recall is $\mathrm{TP}/(\mathrm{TP}+\mathrm{FN})$ over exact matches, and F1 is their harmonic mean. If the input transcript exactly matches the reference sequence, there are no false positives: precision is one, recall equals the fraction $r$ of words marked found, and $F_1=2r/(1+r)$.

Boundary metrics, in contrast, are computed only over exact ASR/reference word matches. Substitutions and insertions are wrong ASR tokens with no valid reference interval, while deletions have no input token for the aligner to place; all three are therefore excluded from boundary scoring.

\section{Text Normalization and the Expand - Merge Contract}
\label{app:textnorm}

\subsection{Why transcript form is a major problem}
An aligner's two inputs are audio and text, and the literature scrutinizes only the first. The text is assumed to be a verbatim spoken-form transcription, because that is what phonetics corpora usually contain e.g. ``twenty eight'' instead of ``28''.

In real-world audios, ASR systems may apply \emph{inverse} text normalization to make output readable, turning ``two thousand one'' into ``2001'' and ``three dollars and fourteen cents'' into ``\$3.14''. Users then edit that readable text, and it is the edited text that is handed back for alignment. So the production transcript is written-form while the audio remains spoken-form, and the mismatch is not merely lexical - it is also a \emph{cardinality} mismatch. One written word token can correspond to one spoken word (e.g. ``cat'') or more (e.g. ``2001'' $\to$ ``two thousand one'' has three). Any interface that assumes one interval per whitespace token, which every aligner API does, is therefore under-specified on real input.

The consequences differ by aligner family. Lexicon-and-HMM systems look up ``2001'' in a pronunciation dictionary and miss. Gentle and MFA fall back to grapheme-to-phoneme conversion, which produces a spelling-driven pronunciation bearing no relation to what was spoken; the word is then either misplaced or reported as not-found. Nyra instead phonemizes the token with espeak-ng, whose rule-based front-end reads many numerals and abbreviations aloud. CTC aligners work on characters, and digits and symbols fall outside their alphabets. MMS-FA maps such characters to a wildcard token and still emits an interval whose extent is arbitrary, while WhisperX leaves the word untimed. Speech-LLM aligners inherit their tokenizer's handling of digits, which is notoriously inconsistent.

None of this is measurable on the corpora the field uses, because those corpora contain essentially no written-form tokens. This is a concrete instance of the paper's broader claim: the benchmark determines which failure modes are visible, and an entire class of production failures has been invisible.

\subsection{The expand - merge contract}
Let the transcript be tokens $t_1, \dots, t_N$ - the units the user sees and edits, and the units in which results must be reported. Define an expansion $\phi$ mapping each token to a non-empty sequence of spoken-form words. Concatenating gives $u_1, \dots, u_M$ with $M \ge N$, together with contiguous half-open index spans $[\ell_k, h_k)$ such that $\phi(t_k) = u_{\ell_k}, \dots, u_{h_k - 1}$, $\ell_1 = 1$, $h_N = M+1$, and $\ell_{k+1} = h_k$.

The aligner is run on $u_1, \dots, u_M$, returning intervals $(\hat{s}_j, \hat{e}_j)$ with the null interval $\nullint$ for words it could not place. Results are then \emph{merged} back onto the original tokens. Writing $D_k = \{ j \in [\ell_k, h_k) : \hat{s}_j \neq \varnothing_s \}$ for the located words of group $k$:
\[
(\hat{s}^{\text{tok}}_k, \hat{e}^{\text{tok}}_k) =
\begin{cases}
\bigl(\hat{s}_{\min D_k},\; \hat{e}_{\max D_k}\bigr) & D_k \neq \emptyset\\
\nullint & D_k = \emptyset .
\end{cases}
\]
That is: a token spans from the start of its first located word (lowest index in $D_k$) to the end of its last located word (highest index), and a token whose entire expansion went unlocated is reported as absent - so it flows into the detection metrics and the uniform fallback (Section~\ref{sec:coverage}) exactly like any other unaligned word, rather than being silently assigned a fabricated interval.

\section{Benchmark Construction Details}
\label{app:benchmark-details}

Construction begins with a private collection of real-world recordings that defines the target distribution. We compute only aggregate per-axis distributions from these recordings and do not release them. Candidate clips from the public corpora are tagged with the same measurements, filtered for eligibility, and selected by stratified sampling so that \bench{} approximates the target distribution across the characteristics below. Selection also limits repeated speakers and recordings.

\textbf{Duration.}
We match both coarse duration bins and duration quantiles so \bench{} includes the same balance of short and long recordings as the target distribution.

\textbf{Audio quality.}
Each clip receives a categorical recording-condition label. Public clips are stratified to reproduce the balance of clean, noisy, telephone, clipped, and reverberant conditions observed in the target distribution.

\textbf{Active level.}
Speech-active signal level is bucketed with common thresholds and matched to preserve variation in recording gain.

\textbf{Signal-to-noise ratio.}
A VAD-based speech/non-speech power estimate defines SNR strata, which are matched between the target and public distributions.

\textbf{Reverberation.}
Estimated energy decay after speech offsets provides an RT60 proxy. Public clips are stratified using the same bins to preserve the target distribution of room effects.

\textbf{Bandwidth.}
Spectral rolloff separates narrowband and wideband recordings. Matching this distribution retains telephone-like and bandwidth-limited speech.

\textbf{Speaker count.}
Diarization assigns clips to single-, two-, or multi-speaker strata. Sampling preserves the corresponding distribution while avoiding repeated speakers where metadata permits.

\textbf{Speaking rate.}
Articulation rate is measured over detected speech rather than full clip duration and bucketed consistently in both pools. Public clips are selected to match the resulting pace distribution.

\textbf{Disfluency.}
Hesitations, repetitions, and false starts are normalized by transcript length and grouped into common bins. This prevents \bench{} from containing only fluent read speech.

\textbf{Text difficulty.}
Written-form features such as numerals, symbols, and entities define text-difficulty strata. Matching these strata preserves variation in the normalization burden presented to an aligner.

\textbf{Accent.}
A shared accent classifier\citep{zuluaga2023commonaccent} assigns broad accent categories to both pools. Sampling targets the private collection's category distribution subject to the accents available in the public corpora.

The sampler balances these objectives jointly, so matching is approximate when the public corpora do not contain a requested combination of properties. The selected clips draw from People's Speech, VoxPopuli, and AMI and span the targeted acoustic, speaker, and transcript conditions. Each source transcript is corrected to verbatim speech, including hesitations and disfluencies, before word-boundary annotation.

\subsection{Licensing and attribution}
\label{app:licensing}
\bench{} is a collection of excerpts from three public corpora together with new annotations: 35 clips from People's Speech, 27 from VoxPopuli, and 11 from AMI. All source audio is redistributable (item-level CC BY 2.0 - 4.0, CC0 1.0, and CC BY 4.0, respectively). The audio is not relicensed: every clip remains under the license of its source. For each clip the released manifest records the source corpus, the source recording identifier and URL, the creator or rights holder and license exactly as stated by the source, the excerpt offsets within the source recording, and the modifications we made (excerpting, conversion to 16\,kHz mono, and correction of the source transcript to verbatim speech).
\begin{itemize}[leftmargin=*, itemsep=1pt]
  \item \textbf{People's Speech} \citep{galvez2021peoplesspeech}, 35 clips. Licensed per item under CC BY 2.0, 2.5, 3.0, or 4.0; we selected only CC BY recordings and excluded those under CC BY-SA. Creator and license URI are given per clip in the manifest.
  \item \textbf{VoxPopuli} \citep{wang2021voxpopuli}, 27 clips. CC0 1.0. The audio originates from European Parliament plenary sessions, which we acknowledge as the original source.
  \item \textbf{AMI Meeting Corpus} \citep{carletta2005ami}, 11 clips. CC BY 4.0, University of Edinburgh.
\end{itemize}
The verbatim transcripts, word boundaries, ASR outputs, selection metadata, and evaluation code that we contribute are released under CC BY 4.0. Anyone redistributing the audio must satisfy the attribution terms of the source licenses; the manifest contains the information needed to do so. TIMIT and Buckeye are used for evaluation only and are not redistributed.

\subsection{Annotation and inter-annotator agreement}
\label{app:annotation}

\textbf{Protocol.}
Before annotating the benchmark, the three annotators completed a calibration round in which they received feedback on their boundary placement, so that they shared one convention for where a word starts and ends. They then independently placed the start and end of every word in every clip, so each boundary carries three independent annotations.

\begin{figure}[ht]
\centering
\includegraphics[width=0.75\textwidth]{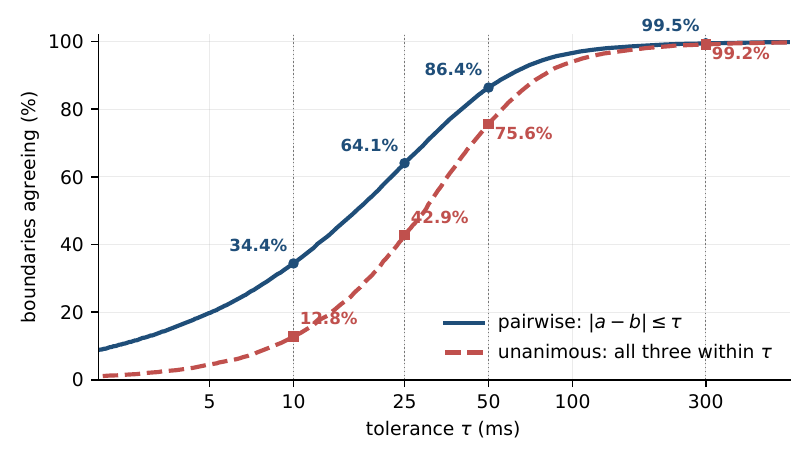}
\caption{\textbf{Inter-annotator agreement on \bench{} word boundaries.} Share of boundaries agreeing within a tolerance $\tau$: pairwise, pooled over the three annotator pairs (solid), and unanimous, all three marks within $\tau$ of each other (dashed). Markers give the values at the reported tolerances.}
\label{fig:rater-agreement}
\end{figure}

\textbf{Agreement.}
Figure~\ref{fig:rater-agreement} plots, as a function of the tolerance $\tau$, the share of boundaries on which annotators agree. The pairwise curve pools the three annotator pairs; the unanimous curve requires all three marks to lie within $\tau$ of one another. Pairwise agreement is 34.4\% at 10\,ms, 64.1\% at 25\,ms, 86.4\% at 50\,ms, and 99.5\% at 300\,ms; unanimous agreement is 12.8\%, 42.9\%, 75.6\%, and 99.2\%. The median pairwise difference is 16.8\,ms and each annotator's robust standard deviation about the others is about 22\,ms, with no difference between annotators. At 10\,ms, humans disagree on two-thirds of boundary pairs, so a 10\,ms tolerance measures annotation noise; 300\,ms is far outside human disagreement, so the catastrophic rate is reliable.

\textbf{Combining the three annotations.}
The reference boundary is a per-boundary median of the three marks, with one structural step taken first. Consecutive words usually share a boundary: annotators mark the end of one word and the start of the next as a single time in 73 - 83\% of junctions. Taking the median of the three ends and, separately, the median of the three starts treats that shared time as two unknowns. Whenever two annotators share a boundary and the third leaves a gap, the two medians are drawn from different marks and generally differ, opening a gap that no annotator marked. On \bench{}, independent medians open 628 such phantom gaps (median width 21\,ms, 90th percentile 61\,ms) and create 148 phantom overlaps, all at junctions where two of three annotators had agreed the words touch. We therefore first decide, at each junction, whether the two words touch by majority vote: they touch if at least two annotators left an absolute gap of at most 5\,ms. A touching junction becomes one boundary, set to the median of the three annotators' junction centres $(e_i+s_{i+1})/2$. Every other boundary, the first start, the last end, and both sides of a non-touching junction, is the median of the three marks. If the medians leave a gap it is kept, and if they leave an overlap it is also kept: words do overlap in \bench{} (crosstalk/interruptions), and 52 of 8{,}000 junctions in the reference overlap by more than 5\,ms. Forcing the reference to be monotonic would misplace exactly those words. The asymmetric error does not charge a prediction for the overlapping part of a neighbor (Appendix~\ref{app:metrics}), so an exact prediction at these junctions scores zero.

\section{Buckeye Reference-Corpus Selection and Preparation}
\label{app:reference-preparation}

Buckeye is distributed as 255 roughly ten-minute interview sessions rather than utterance clips \citep{pitt2007buckeye}. Prior work reports large swings in MFA error on Buckeye and attributes them in part to its multi-minute inputs \citep{rousso2024tradition}. Duration is only part of the problem: although the target speaker wears a microphone up close, the recordings contain interviewer turns while transcript contains only the interviewee's speech. If these regions are retained, the waveform contains speech for which the purported ground truth has no corresponding words or boundaries; errors there reflect an invalid reference - audio pairing rather than aligner accuracy. We therefore evaluate only regions that can be made reliably single-speaker.

Silence-based chunking cannot enforce this condition: a brief, low-level interjection need not be surrounded by a pause, and a silence threshold can join two interviewee turns across intervening speech. We instead use Buckeye's metadata and transcript tags to cut each session into interviewee-only chunks, then apply duration and speaker-diarization filters and resample the audio. This preparation is distinct from the model-window chunking and stitching in Appendix~\ref{app:long-form}: every aligner receives the same prepared samples. After a short-clip, word-count, and single-speaker filter, 6{,}185 clips (244{,}408 words) remain for every reported Buckeye number.

\textbf{Comparison with the MFA 2026 Buckeye benchmark.}
Our Buckeye results are not directly comparable to the lower MFA errors reported by \citet{mcauliffe2026montreal}, because the evaluation units and retained speech differ. That work constructs a 17.12-hour subset by splitting at silences of at least 300\,ms, adding 200\,ms of context on either side, and discarding utterances with three words or fewer or with unknown, cutoff, or excised words. In contrast, our 22.31-hour subset splits only at annotated interviewer, overlap, third-speaker, and defective regions; it deliberately retains internal silence and realized cutoff, hesitation, and error-marked words, and does not impose a maximum turn duration. Consequently, MFA is repeatedly reset on short, transcription-clean utterances in their benchmark, whereas here it must align complete interviewee turns, over which a local pronunciation or transcript mismatch can affect a longer globally constrained path. The system configurations and scoring conventions also differ: their strongest Buckeye word result uses the US-English ARPAbet model and an onset-oriented boundary metric, while our MFA baseline uses the public Global-English model bundle and our symmetric comparison pools both starts and ends. The absolute MFA values therefore reflect different subsets, segmentation regimes, model bundles, and boundary sets rather than a replication of the same condition.

\textbf{Controlled re-segmentation experiment.}
To isolate the effect of evaluation units, we reconstruct the published preprocessing rule from the raw Buckeye tiers and rerun both MFA and \sys{} with ground-truth transcripts. Because the released MFA benchmark code begins from an already-segmented database and provides neither its upstream importer nor an utterance manifest, exact identity is not possible. We interpret a silence as any interval of at least 300\,ms without a lexical reference token, add 200\,ms of real neighboring audio on each side, and reject utterances with at most three words or an unknown, cutoff, or excised token. This produces 20{,}232 utterances, 230{,}821 words, and 18.05 hours, close to the reported 17.12 hours. Every system receives these same clips and the same normalized ground-truth transcript.

\begin{table}[ht]
\centering
\small
\caption{\textbf{MFA-style Buckeye utterances.} Both systems are rerun on our reconstruction of the segmentation used by \citet{mcauliffe2026montreal}. Mean asymmetric and symmetric errors pool the two boundaries of every scored word; symmetric error retains displacement within silence. A threshold failure means the maximum asymmetric error over a word's start and end exceeds the stated tolerance. All values use ground-truth transcripts and all-word scoring.}
\label{tab:buckeye-mfa-chunks}
\begin{tabular}{lrrrrr}
\toprule
% Generated by eval/scripts/buckeye_mfa_chunks_table.py; do not edit by hand.
% source: /data/mithilesh/aligner/datasets/buckeye_mfa_chunks/error_metrics.csv
% MFA      <- mfa (230807 words)
% \sys{}   <- 136-baseline-fixed_warmup-fixed_lambda_100000 (230807 words)
System & Mean asym.\ (ms) $\downarrow$ & Mean sym.\ (ms) $\downarrow$ & \% ${>}25$\,ms $\downarrow$ & \% ${>}50$\,ms $\downarrow$ & \% ${>}300$\,ms $\downarrow$ \\
\midrule
MFA & 24.75 & 28.83 & \textbf{44.70} & 21.22 & 0.89 \\
\sys{} & \textbf{19.38} & \textbf{26.65} & 47.81 & \textbf{8.36} & \textbf{0.20} \\
\bottomrule

\end{tabular}
\end{table}

The controlled run confirms that segmentation explains much of the discrepancy. With the same Global-English MFA baseline used in our main comparison, mean symmetric error falls from 42.27\,ms on interviewee turns to 28.83\,ms on the short utterances, substantially narrowing the gap to the 25.35\,ms Global-English result reported by \citet{mcauliffe2026montreal}. The residual difference is consistent with the non-identical retained subset, software path, and scoring convention; their stronger 21.75\,ms result additionally uses the US-English ARPAbet model. Under these identical reconstructed inputs (Table~\ref{tab:buckeye-mfa-chunks}), \sys{} still has lower mean asymmetric and symmetric error than MFA and about $4\times$ fewer catastrophic errors (0.20\% versus 0.89\%), while MFA remains better at the 25\,ms tolerance (44.70\% versus 47.81\%).

\section{\sys{} Implementation and Training Details}
\label{app:model-details}

\subsection{Development set and model selection}
\label{app:devset}
Every design decision (architecture, losses, augmentation, label filtering and correction, presence cutoff, evaluation protocol, training length) was made on a held-out private set of manually annotated recordings disjoint from all evaluation data. \bench{}, TIMIT, and Buckeye were used only to evaluate the fixed recipe; the \bench{} ablations and training curves reported in this paper were computed afterwards and informed no decision.

\subsection{Architecture}
At 16\,kHz, a strided convolution uses a 512-sample window and 256-sample hop. Dia character features retain a one-to-one token map; character offsets identify each word span, whose contextualized vectors are mean-pooled. Audio and text receive modality embeddings and pass through 12 joint self-attention layers (width 1024, 16 heads, FFN width 4096). Audio RoPE positions are the integer frame indices $0,\dots,L-1$. Text token $k\in\{0,\dots,C-1\}$ receives the continuous position $\frac{k}{C-1}\,L$, where $L$ is the number of valid (unpadded) audio frames of that sample. Approximately 201M of the 213M trainable parameters are in this transformer. The frozen Dia text encoder adds 252M parameters (we load only the encoder of the 1.6B-parameter checkpoint), so the full inference-time model has about 465M parameters. Ablation 5 in Table~\ref{tab:model-ablations} replaces Dia with learned character embeddings; it is slightly worse on every metric, but the differences lie within the marginal intervals, so the ablation does not establish a clear advantage for either representation.

The refinement head first projects the 1024-wide audio outputs to 256 channels with a $1{\times}1$ convolution, then applies four transposed-convolution stages (kernel 4, stride 2, padding 1, 256 channels, GELU), each doubling the frame rate, to go from the 16\,ms patch grid to 1\,ms. A final convolution (kernel 5, padding 2, GELU) refines the 1\,ms features and a $1{\times}1$ projection produces the 256-dimensional per-frame features $y_s$ and $y_e$; the start and end branches have separate weights. Separate projections turn each pooled word into start/end classifiers, whose dot products with millisecond audio features produce $N\times T_{\mathrm{ms}}$ heatmaps. Decoding independently selects peaks for start/end. The presence head (Section~\ref{sec:presence-head}) is a two-layer projection of the pooled word vector with a scalar bias initialized to $+2$ (a prior that words are spoken); a word is declined and emitted as $\nullint$ when its probability is below 0.1. Training inserts transcript words at a 5\% rate as negative examples, drawing each fake word's surface from the same chunk so that the head cannot solve the task from vocabulary alone.

\subsection{Long-form inference}
\label{app:long-form}
The model accepts at most $L=30$\,s of audio at a time. We therefore process a shorter recording in one call and split a longer recording into chunks. Each chunk contains a central region plus up to $p=2$\,s of context on either side. We prefer to split at silences lasting at least $g=0.5$\,s. To estimate where transcript words occur, the caller may provide approximate ASR word times; otherwise, we run \whisper{} once on the full recording and cache its output. These times are used only to choose chunks and route words to them. The final word boundaries always come from \sys{}.

We match the ASR transcript to the target words $w_{1:N}$ in order using \texttt{SequenceMatcher}. Before matching, we lowercase words and remove non-alphanumeric characters. Only exact matches are used as candidate anchors. We reject a candidate if its timestamp jumps implausibly far forward from the previous accepted anchor. Specifically, a candidate with ASR start $s_i$ is kept only if
\[
s_i-e_j \leq 10+0.5k,
\]
where $e_j$ is the end of the previous accepted anchor and $k$ is the number of target words between the two anchors. The 10\,s allowance permits a real pause; the additional $0.5k$\,s permits a larger gap when the ASR skipped words. This check prevents a repeated phrase from matching to a distant occurrence. We estimate times for unmatched words by interpolating between accepted anchors. Unmatched words at the beginning or end are spread over the remaining ASR time range instead of being assigned the same time. Finally, we force start and end times to be non-decreasing and ensure that every end is at or after its start.

We discard this approximate word-time map if fewer than two anchors survive (one for a transcript of at most two words), or if fewer than half of the candidate anchors survive the jump check. When this happens, we still divide the transcript among the chunks in order. A chunk receives words in proportion to how much detected speech it contains, with word character counts used as a rough estimate of speaking time. This fallback avoids positioning a whole recording from only one unreliable match.

To choose the audio cuts, Silero VAD \citep{silero2024vad} first identifies speech and silence. We prefer gaps of at least $g$ seconds and rank them using the approximate word times: a silence between words is preferred to one that appears to cut through a word. Because a model input may include $p$ seconds of context on each side, the central core can be at most
\[
C=L-2p=26\,\mathrm{s}.
\]
Moving from left to right, we choose the latest suitable cut within this 26\,s budget. We use a hard cut only if no silence or word boundary is available. The result is a sequence of ordered, non-overlapping cores $[a_r,b_r)$.

When the approximate word-time map is available, each word is assigned to one core. The boundary between the word ranges of consecutive cores is the midpoint $(b_r+a_{r+1})/2$, and assignment uses each word's approximate start time. The audio sent to the model initially covers the core plus context,
\[
[\max(0,a_r-p),\min(T,b_r+p)].
\]
Its transcript includes both the words owned by the core and any additional words whose approximate times fall in the context.

We align each chunk independently and add the chunk's sample-exact offset to convert its predictions back to full-recording time. A word near a cut may be predicted by more than one chunk. We first prefer a prediction whose center $(\hat{s}_i+\hat{e}_i)/2$ falls inside that chunk's core. If several do, we choose the one farthest from a core edge, where the model had the most context. If none falls inside a core, we choose the prediction nearest to one. Thus every target word appears once in the output, and all final boundaries come from \sys{}, not from the coarse ASR timestamps.

\subsection{Loss}
\label{app:loss}
Equations~\ref{eq:focal} and \ref{eq:argmax} are applied to the start and end heatmaps separately. Each boundary target is a 1\,ms-grid Gaussian with $\sigma=10$\,ms, truncated at $4\sigma$. The focal loss of \citet{lin2017focal} is defined for binary labels; we extend it to the soft target $g\in[0,1]$ by interpolating both the probability of the target class and the class weight linearly in $g$:
\begin{gather*}
\mathrm{FocalBCE}(p,g)=\alpha_g\,(1-p_g)^{\gamma}\,\mathrm{BCE}(p,g),\\
p_g = p\,g+(1-p)(1-g),\qquad
\alpha_g=\alpha\,g+(1-\alpha)(1-g),
\end{gather*}
where $\mathrm{BCE}(p,g)=-g\log p-(1-g)\log(1-p)$ and $1-p_g$ is clamped below at $10^{-6}$. For $g\in\{0,1\}$ this is the standard focal loss. We use $\alpha=0.25$ and $\gamma=2$, normalize by the target mass $\sum_{k,t}g_{k,t}$, making the term independent of clip length, and assign it weight $\lambda_{\mathrm{f}}=1$.

The soft-argmax displacement is divided by the clip length $T$, measured on the 1\,ms grid over the valid (unpadded) frames of each sample. Its penalty is the smooth-$L_1$ function
\[
\ell(x)=\operatorname{SmoothL1}_{\beta}(x)=
\begin{cases}
x^2/(2\beta) & |x|<\beta,\\
|x|-\beta/2 & \text{otherwise,}
\end{cases}
\qquad \beta=0.01,
\]
weighted by $\lambda_{\mathrm{a}}=10$. Because the displacement is expressed relative to $T$, the quadratic-to-linear transition occurs at an absolute error of $\beta T$: 300\,ms for a 30\,s clip but 100\,ms for a 10\,s clip, and the gradient of the linear regime is likewise $\lambda_{\mathrm{a}}/T$ per unit time.

Presence uses $\mathcal{L}_{\mathrm{pres}}=\tfrac12\big[\overline{\mathrm{BCE}}_{y_k=1}+\overline{\mathrm{BCE}}_{y_k=0}\big]$ over the word probabilities $p_k$ and labels $y_k\in\{0,1\}$, where inserted words have $y_k=0$. Each bar denotes the mean for one class in the batch. Positives and negatives therefore carry equal total weight even though inserted words make up only ${\sim}5\%$ of the transcript. We set $\lambda_{\mathrm{p}}=0.1$. The total loss is $\lambda_{\mathrm{f}}(\mathcal{L}^{s}_{\mathrm{focal}}+\mathcal{L}^{e}_{\mathrm{focal}})+\lambda_{\mathrm{a}}(\mathcal{L}^{s}_{\mathrm{argmax}}+\mathcal{L}^{e}_{\mathrm{argmax}})+\lambda_{\mathrm{p}}\mathcal{L}_{\mathrm{pres}}$.

\subsection{Collapsed pseudo-label filtering}
\label{app:pseudo-label-filtering}
Gentle occasionally assigns an ordinary word a one-to-three-frame interval. We round each pseudo-label duration to milliseconds and mask its boundary loss when it is shorter than 30\,ms, except for a plausible short function word: at most three characters and surrounded on both sides by words whose adjacent gaps are at most 10\,ms (overlap counts as touching). This exception retains labels such as ``a,'' ``I,'' and ``the'' in continuous speech while preventing suspicious collapsed labels from supervising the boundary heads. The word remains in the text context; only its boundary loss is masked.

\subsection{Augmentation}
A master augmentation gate selects 25\% of training samples; conditional on that gate, independent transforms may be composed. Synthetic echo/reverberation is applied with probability 0.40 using an exponentially decaying room impulse response with RT60 sampled uniformly from 0.1 - 1.0\,s and direct-to-reverberant ratio from 0 - 25\,dB. Background noise is applied with probability 0.40 by mixing a random crop from a 512-item, 10\,s AudioSet pool at 3 - 20\,dB SNR. Hard clipping is applied with probability 0.10 at a symmetric clipped fraction sampled from 0 - 0.10. Low-pass filtering has probability 0.30 with cutoff 2 - 5\,kHz; high-pass filtering has probability 0.15 with cutoff 50 - 100\,Hz; and gain perturbation has probability 0.40 over $-12$ to $+12$\,dB. Ablation 7 disables this entire waveform chain but retains the rest of the training recipe. Synthetic transcript insertions, sampled at 5\%, separately supervise word presence.

\subsection{EMA label correction and training dynamics}
\label{app:training-curve}
\textbf{Optimization.}
Training runs 100,000 optimizer steps at batch size 128 with AdamW \citep{loshchilov2019decoupled}, $\beta=(0.9,0.95)$, $\epsilon=10^{-8}$, peak learning rate $3\times10^{-4}$, and weight decay 0.03 applied to weight matrices only (norms, biases, and embeddings are excluded). The learning rate warms up linearly over the first 2,000 steps and then follows a cosine decay to zero at 100k steps.

\textbf{Label correction.}
Error on a held-out split of the pseudo-labeled training corpus continues to decrease throughout training and therefore does not expose overfitting to incorrect Gentle labels. In contrast, error on our held-out private development set of manually annotated recordings is lowest around 20k steps and rises thereafter. This discrepancy motivates the correction procedure: a rolling bank of three EMA snapshots begins at 20k steps, before the development error degrades. The development set is disjoint from \bench{}, TIMIT, and Buckeye, and the correction start, EMA decay, bank size, and agreement thresholds were all chosen on it. The EMA weights track the online model with decay 0.999 per optimizer step. A snapshot of the EMA weights is copied into the bank every 10k steps, starting at 20k, so the bank is full from 40k steps and thereafter always holds the three most recent snapshots. Labels within 50\,ms of every teacher retain Gentle; if a full bank conflicts with Gentle but agrees internally within 20\,ms, the ensemble median replaces it; conflicts without consensus are masked.

Table~\ref{tab:training-curve} evaluates every saved checkpoint of the baseline and of the same recipe with EMA-snapshot correction disabled (ablation 9 of Table~\ref{tab:model-ablations}) on \bench{} with the hand-corrected transcript. We computed this table after both runs had finished; it illustrates the effect of correction on held-out data and played no role in choosing the correction schedule. Checkpoints are saved every 10k steps and indexed by the hours of audio sampled so far, computed as steps $\times$ batch size 128 $\times$ a mean clip length of 22\,s, so 10k steps is roughly 7{,}800\,h and 100k steps roughly 78{,}000\,h. Correction starts at 20k steps (${\sim}15{,}600$\,h); the rule in the table follows the last checkpoint trained without it.

Before correction starts, the two runs are comparable (0.43\% versus 0.45\% catastrophic words at 20k). They then diverge. The corrected run's mean falls from 20.01\,ms at 30k to about 17.3\,ms from 50k onward and ends at 17.16\,ms, whereas the uncorrected run reaches its lowest mean at 50k (17.87\,ms) and drifts back up to 19.46\,ms. The catastrophic tail separates more clearly: the uncorrected run's ${>}300$\,ms rate rises from 0.45\% at 20k to 0.68\% at 100k, while the corrected run stays between 0.35\% and 0.43\% and ends at 0.38\%. At 100k, disabling correction therefore produces about 1.8 times as many catastrophic words and a 2.3\,ms higher mean. EMA-snapshot correction primarily prevents late-training degradation, which is most visible in the tail.

\begin{table}[ht]
\centering
\small
\caption{Training dynamics on \bench{} with the hand-corrected transcript at every saved checkpoint, with EMA-snapshot label correction on (baseline) and off (ablation 9). We report both mean asymmetric error and the percentage of words with error above 300\,ms. Rows are the hours of audio sampled so far; the rule marks where correction starts at 20k steps. Each difference is off minus on, and the best checkpoint in each on/off column is bold.}
\label{tab:training-curve}
\begin{tabular}{rrrrrrr}
\toprule
& \multicolumn{3}{c}{Mean asym.\ error (ms) $\downarrow$} & \multicolumn{3}{c}{Words ${>}300$\,ms (\%) $\downarrow$} \\
\cmidrule(lr){2-4}\cmidrule(lr){5-7}
Audio sampled (h) & EMA on & EMA off & Difference & EMA on & EMA off & Difference \\
\midrule
7{,}822 & 26.46 & 24.53 & -1.93 & 0.84 & 0.88 & +0.04 \\
15{,}644 & 18.52 & 19.43 & +0.91 & 0.43 & \textbf{0.45} & +0.01 \\
\midrule
23{,}467 & 20.01 & 18.64 & -1.38 & 0.38 & 0.47 & +0.09 \\
31{,}289 & 18.33 & 19.63 & +1.30 & 0.41 & 0.52 & +0.11 \\
39{,}111 & 17.43 & \textbf{17.87} & +0.44 & \textbf{0.35} & 0.51 & +0.16 \\
46{,}933 & 17.37 & 19.14 & +1.77 & 0.43 & 0.58 & +0.15 \\
54{,}756 & 17.29 & 19.28 & +1.99 & 0.37 & 0.61 & +0.24 \\
62{,}578 & 17.33 & 19.49 & +2.15 & 0.42 & 0.69 & +0.27 \\
70{,}400 & 17.17 & 19.40 & +2.23 & 0.37 & 0.66 & +0.28 \\
78{,}222 & \textbf{17.16} & 19.46 & +2.30 & 0.38 & 0.68 & +0.30 \\
\bottomrule

\end{tabular}
\end{table}

\section{Additional Ablations and Analyses}

\subsection{Protocol Ablations}
\label{app:conditions}
Table~\ref{tab:conditions} changes one protocol choice at a time from the default of Table~\ref{tab:development-results} and reports each system's default score followed by the delta. \emph{Chunking} re-runs a baseline on the context-padded chunks and word lists \sys{} used, stitched by the same ownership rule (Appendix~\ref{app:long-form}). \emph{No text normalization} feeds written forms such as ``2001'' verbatim (Section~\ref{sec:textnorm}). \emph{Aligned-only scoring} drops the uniform fallback and scores only the words a system placed. \sys{} always chunks, so its baseline is already chunked and has no separate chunking row.

\begin{table}[ht]
\centering
\scriptsize
\setlength{\tabcolsep}{3pt}
\caption{Protocol conditions on \bench{} with the hand-corrected transcript. \sys{} always chunks, so its baseline is already chunked and it has no chunking row. Error-rate changes are coloured when their magnitude is at least 20\% of the system's baseline rate; mean-error changes are coloured at 2\,ms and F1 changes at 0.5 points. \better{Green} improves and \worse{red} degrades.}
\label{tab:conditions}
\begin{tabular*}{\textwidth}{@{\extracolsep{\fill}}llrrrrr@{}}
\toprule
Protocol & System & \% ${>}300$\,ms $\downarrow$ & \% ${>}50$\,ms $\downarrow$ & \% ${>}25$\,ms $\downarrow$ & Mean asym.\ (ms) $\downarrow$ & F1 (\%) $\uparrow$ \\
\midrule
baseline & \sys{} & 0.38 & 7.27 & 29.82 & 17.16 & 99.82 \\
 & Gentle & 2.11 & 12.61 & 35.20 & 31.43 & 98.90 \\
 & MFA & 4.15 & 34.16 & 64.92 & 73.00 & 100.00 \\
 & Nyra & 1.28 & 29.99 & 68.84 & 33.18 & 100.00 \\
 & MMS-FA & 2.49 & 46.69 & 80.74 & 48.91 & 98.96 \\
 & WhisperX & 1.41 & 53.25 & 84.75 & 49.04 & 99.78 \\
\midrule
chunking & Gentle & \worse{+0.55} & +0.71 & +0.84 & \worse{+69.72} & -0.15 \\
 & MFA & -0.31 & -0.67 & -0.17 & \better{-10.69} & +0.00 \\
 & Nyra & +0.01 & \better{-10.44} & -13.10 & -1.82 & +0.00 \\
 & MMS-FA & -0.37 & -0.63 & +0.08 & \worse{+3.29} & \better{+1.03} \\
 & WhisperX & +0.03 & +0.69 & +0.53 & \worse{+8.31} & +0.22 \\
\midrule
no text normalization & \sys{} & +0.00 & -0.04 & -0.09 & -0.04 & +0.01 \\
 & Gentle & +0.07 & +0.09 & -0.13 & +0.38 & +0.07 \\
 & MFA & \worse{+1.09} & +0.74 & +0.33 & \worse{+9.23} & \worse{-1.11} \\
 & Nyra & +0.12 & +0.00 & -0.01 & +1.51 & -0.11 \\
 & MMS-FA & +0.06 & +0.00 & -0.12 & +0.79 & \worse{-0.52} \\
 & WhisperX & \worse{+0.53} & +0.43 & +0.12 & \worse{+13.68} & -0.06 \\
\midrule
aligned-only scoring & \sys{} & -0.05 & -0.09 & -0.08 & -0.16 & +0.00 \\
 & Gentle & \better{-0.59} & -1.06 & -0.94 & \better{-6.99} & +0.00 \\
 & MFA & +0.00 & +0.00 & +0.00 & +0.00 & +0.00 \\
 & Nyra & +0.00 & +0.00 & +0.00 & +0.00 & +0.00 \\
 & MMS-FA & -0.15 & -0.16 & +0.01 & -0.49 & +0.00 \\
 & WhisperX & -0.08 & -0.11 & -0.02 & -0.27 & +0.00 \\
\bottomrule

\end{tabular*}
\end{table}

\textbf{Chunking.} Only MFA and Nyra benefit from chunking. Both commit to a single globally constrained Viterbi path over the whole clip, with no mechanism to repair a locally bad alignment. MFA's whole-clip error is already high (73.00\,ms mean) as local mismatches propagate to later words. Shorter chunks limit how far such errors can cascade, reducing MFA's mean error by 10.69\,ms and its catastrophic rate by 0.31 points. Nyra gains in the moderate range instead: its rate above 50\,ms falls by 10.44 points (29.99\% to 19.55\%) and its rate above 25\,ms by 13.10 points, while its catastrophic rate is unchanged and its mean falls by 1.82\,ms. This pattern suggests that its boundaries drift on long recordings rather than failing outright, and that chunking bounds the drift. Even on \sys{}'s chunks, Nyra remains well behind \sys{} (19.55\% vs.\ 7.27\% above 50\,ms). The other baselines already handle whole clips well, so chunking only adds errors at chunk seams: Gentle's mean error rises by 69.72\,ms and its catastrophic rate by 0.55 points, WhisperX's mean by 8.31\,ms, and MMS-FA's by 3.29\,ms.

\textbf{Text normalization.} Both MFA and WhisperX are affected by text normalization. With a limited pronunciation dictionary and no fallback for words outside it, unnormalized written forms raise MFA's mean error by 9.23\,ms, its catastrophic rate by 1.09 points, and lower its F1 by 1.11 points. WhisperX is also affected (+13.68\,ms mean, +0.53 points catastrophic) because its English aligner is a letter-level wav2vec\,2.0 CTC model whose vocabulary is alphabetic; unnormalized digits and symbols fall outside that vocabulary or match the acoustics poorly. MMS-FA, trained on large amounts of character-level data, is robust (+0.79\,ms), as is \sys{} ($-0.04$\,ms); Gentle re-decodes unmatched regions and changes little (+0.38\,ms). Nyra, which phonemizes OOV tokens with espeak-ng, moves by $+1.51$\,ms.

\textbf{Aligned-only scoring.} Aligned-only scoring lowers Gentle's catastrophic rate by 0.59 points and its mean error by 6.99\,ms, because the words it rejects are exactly the ones the fallback scores against it. The other systems place almost every word, so excluding unaligned words changes their scores by less than 0.6\,ms. Nyra places every word (F1 $100\%$), so aligned-only scoring is identical to the headline protocol. This supports the all-word fallback from Section~\ref{sec:coverage}, which was fixed before any system was run on the evaluation data, as the headline protocol.

\subsection{Performance by \bench{} Subgroup}
\label{app:subgroup-results}

Table~\ref{tab:subgroup-results} splits the hand-corrected \bench{} results by accent, speaker count, and written-word difficulty using the same fallback and metrics as Table~\ref{tab:development-results}. These are descriptive rather than controlled comparisons: accent and speaker count covary with source, acoustics, and content.

\textbf{Accent.}
Accent effects are mixed across metrics. Several systems have lower mean error on the non-American slice while simultaneously having more catastrophic errors, and MFA's mean worsens substantially. For multilingual media pipelines this makes accent an important audit dimension, but the 1{,}701-word slice also differs in source and recording conditions, so these data support no causal claim about accent handling.

\textbf{Speaker count.}
Multi-speaker clips are the clearest difficult slice: every system has more catastrophic errors and higher mean error than on single-speaker clips. MFA has the largest increase, while \sys{} remains best on both sides. This is directly relevant to meetings, podcasts, and interviews: speaker changes and overlap should trigger extra quality control even when the transcript is correct.

\textbf{Word type.}
Difficult written forms widen the 50\,ms error rate most for Gentle and MFA, less for \sys{}, somewhat for Nyra (29.4\% to 36.2\%), and negligibly for MMS-FA and WhisperX. Numbers, abbreviations, and named entities are common in edited transcripts, so this sensitivity affects exactly the tokens users are likely to search or cut around. Mean changes are mixed; the supported conclusion is additional medium-sized errors in the lexicon-and-HMM systems (Gentle and MFA, and to a lesser degree Nyra), not uniform degradation, while the character-level CTC aligners are largely insensitive.

\begin{table}[ht]
\centering
\scriptsize
\setlength{\tabcolsep}{3pt}
\caption{\textbf{Alignment performance by \bench{} subgroup with hand-corrected transcripts.} Each cell is \emph{easy / difficult}: American/non-American for accent, single-/multi-speaker for speaker count, and ordinary/difficult for word type. The $n$ values give the number of scored words on the two sides. Bold marks the best system and underline the second best separately on each side of each split.}
\label{tab:subgroup-results}
\resizebox{\textwidth}{!}{%
\begin{tabular}{llrrrrrr}
\toprule
Split & Metric & \sys{} & Gentle & MFA & Nyra & MMS-FA & WhisperX \\
\midrule
% Generated by eval/scripts/subgroup_table.py; do not edit by hand.
% accent: 6371 / 1701 words
\multirow{5}{*}{\shortstack{Accent\\$n=6371/1701$}} & \% ${>}300$\,ms $\downarrow$ & \textbf{0.30} / \textbf{0.71} & 2.07 / 2.23 & 3.53 / 6.47 & \underline{1.19} / 1.59 & 2.26 / 3.35 & 1.41 / \underline{1.41} \\
 & \% ${>}50$\,ms $\downarrow$ & \textbf{7.3} / \textbf{7.1} & \underline{12.2} / \underline{14.1} & 34.5 / 32.7 & 30.4 / 28.3 & 47.7 / 42.7 & 54.0 / 50.3 \\
 & \% ${>}25$\,ms $\downarrow$ & \textbf{30.2} / \textbf{28.6} & \underline{34.8} / \underline{36.5} & 65.0 / 64.5 & 68.8 / 69.0 & 81.2 / 79.1 & 85.2 / 83.0 \\
 & Mean asym.\ (ms) $\downarrow$ & \textbf{17.41} / \textbf{16.22} & \underline{32.65} / \underline{26.86} & 66.59 / 97.03 & 32.98 / 33.92 & 48.18 / 51.66 & 49.83 / 46.11 \\
 & F1 (\%) $\uparrow$ & \underline{99.83} / 99.76 & 98.97 / 98.63 & \textbf{100.00} / \textbf{100.00} & \textbf{100.00} / \textbf{100.00} & 98.89 / 99.23 & 99.76 / \underline{99.82} \\
\midrule
% speakers: 4851 / 3221 words
\multirow{5}{*}{\shortstack{Speaker count\\$n=4851/3221$}} & \% ${>}300$\,ms $\downarrow$ & \textbf{0.06} / \textbf{0.87} & 1.20 / 3.48 & 1.79 / 7.70 & 0.80 / \underline{1.99} & 1.61 / 3.82 & \underline{0.78} / 2.36 \\
 & \% ${>}50$\,ms $\downarrow$ & \textbf{5.8} / \textbf{9.6} & \underline{10.3} / \underline{16.1} & 30.9 / 39.1 & 28.5 / 32.2 & 47.3 / 45.8 & 55.0 / 50.6 \\
 & \% ${>}25$\,ms $\downarrow$ & \textbf{29.2} / \textbf{30.8} & \underline{34.2} / \underline{36.8} & 62.7 / 68.2 & 68.0 / 70.1 & 81.2 / 80.0 & 85.7 / 83.4 \\
 & Mean asym.\ (ms) $\downarrow$ & \textbf{13.91} / \textbf{22.04} & \underline{21.06} / 47.06 & 36.13 / 128.53 & 30.47 / \underline{37.26} & 45.50 / 54.06 & 47.09 / 51.99 \\
 & F1 (\%) $\uparrow$ & 99.92 / \underline{99.67} & 99.41 / 98.12 & \textbf{100.00} / \textbf{100.00} & \textbf{100.00} / \textbf{100.00} & 99.14 / 98.69 & \underline{99.94} / 99.53 \\
\midrule
% words: 7353 / 719 words
\multirow{5}{*}{\shortstack{Word type\\$n=7353/719$}} & \% ${>}300$\,ms $\downarrow$ & \textbf{0.38} / \textbf{0.42} & 1.96 / 3.62 & 3.94 / 6.26 & \underline{1.26} / 1.39 & 2.57 / 1.67 & 1.43 / \underline{1.25} \\
 & \% ${>}50$\,ms $\downarrow$ & \textbf{7.1} / \textbf{9.3} & \underline{12.2} / \underline{17.1} & 33.6 / 40.3 & 29.4 / 36.2 & 46.6 / 47.1 & 53.3 / 52.9 \\
 & \% ${>}25$\,ms $\downarrow$ & \textbf{29.8} / \textbf{30.2} & \underline{34.6} / \underline{41.3} & 64.7 / 67.5 & 68.2 / 75.2 & 80.7 / 81.4 & 85.1 / 81.5 \\
 & Mean asym.\ (ms) $\downarrow$ & \textbf{17.27} / \textbf{15.95} & \underline{31.07} / \underline{35.20} & 74.73 / 55.29 & 32.66 / 38.41 & 49.12 / 46.81 & 49.26 / 46.83 \\
 & F1 (\%) $\uparrow$ & \underline{99.85} / 99.51 & 99.36 / 93.95 & \textbf{100.00} / \textbf{100.00} & \textbf{100.00} / \textbf{100.00} & 98.98 / 98.73 & 99.78 / \underline{99.79} \\
\bottomrule

\end{tabular}
}
\end{table}

\subsection{Model Ablations with Noisy Transcripts}
\label{app:model-ablations-asr}

Table~\ref{tab:model-ablations-asr} repeats the model ablations from Table~\ref{tab:model-ablations} under real ASR transcripts. The fine-tolerance error rate changes little across transcript sources, while noisier transcripts primarily increase mean error and catastrophic failures. EMA-snapshot correction remains especially important under \whisper{}, where disabling it produces the largest catastrophic-error rate among the ablations.

\begin{table}[ht]
\centering
\scriptsize
\setlength{\tabcolsep}{3pt}
\caption{\textbf{Model ablations on \bench{} under hand-corrected and noisy transcripts.} Each cell is \emph{hand-corrected / \scribe{} / \whisper{}} for the same audio. Lower is better. ASR conditions score exact ASR - reference word matches only, following Section~\ref{sec:metrics}.}
\label{tab:model-ablations-asr}
\resizebox{\textwidth}{!}{%
\begin{tabular}{llrrrr}
\toprule
Ablation & Change from the baseline & Mean asym.\ (ms) $\downarrow$ & \% ${>}300$\,ms $\downarrow$ & \% ${>}50$\,ms $\downarrow$ & \% ${>}25$\,ms $\downarrow$ \\
\midrule
-- & baseline (${\sim}213$M params) & 17.16 / 20.43 / 32.82 & 0.38 / 0.43 / 1.71 & 7.3 / 7.2 / 7.7 & 29.8 / 29.9 / 29.8 \\
\midrule
1 & width = 512, FFN = 2048 (${\sim}60$M params) & 17.75 / 18.01 / 28.17 & 0.35 / 0.44 / 1.50 & 7.1 / 7.1 / 7.6 & 29.6 / 29.5 / 30.0 \\
2 & 24 layers (${\sim}415$M params) & 17.01 / 17.04 / 30.90 & 0.33 / 0.33 / 1.61 & 7.4 / 7.5 / 7.9 & 30.1 / 30.3 / 30.3 \\
3 & width = 2048, FFN = 8192 (${\sim}821$M params) & 22.23 / 21.44 / 34.05 & 0.33 / 0.35 / 1.50 & 7.3 / 7.2 / 7.7 & 30.0 / 29.8 / 30.0 \\
4 & window = 1024, hop = 512 samples & 17.77 / 17.13 / 29.06 & 0.32 / 0.41 / 1.72 & 7.3 / 7.2 / 7.9 & 29.9 / 29.6 / 30.3 \\
5 & learned character embeddings (instead of frozen Dia) & 17.67 / 18.37 / 27.56 & 0.40 / 0.45 / 1.70 & 7.8 / 7.8 / 8.0 & 30.0 / 29.9 / 30.0 \\
6 & batch size = 64 & 17.21 / 17.36 / 32.55 & 0.37 / 0.43 / 1.51 & 7.6 / 7.5 / 7.8 & 29.8 / 30.0 / 29.8 \\
\midrule
7 & no audio augmentation & 18.05 / 19.24 / 32.09 & 0.32 / 0.29 / 1.72 & 7.7 / 7.7 / 8.1 & 30.5 / 30.5 / 30.4 \\
8 & crosstalk augmentation & 17.98 / 20.82 / 28.07 & 0.35 / 0.40 / 1.60 & 7.5 / 7.6 / 7.8 & 30.0 / 29.8 / 30.0 \\
\midrule
9 & EMA-snapshot correction off & 19.46 / 19.61 / 37.95 & 0.68 / 0.72 / 2.71 & 8.3 / 8.4 / 9.6 & 30.9 / 30.7 / 31.5 \\
\midrule
10 & no soft-argmax loss & 19.83 / 19.52 / 31.88 & 0.38 / 0.36 / 1.53 & 7.7 / 7.6 / 7.9 & 30.1 / 30.1 / 30.1 \\
11 & 16\,ms output grid (no convolutional upsampling) & 21.62 / 24.78 / 33.80 & 0.41 / 0.50 / 1.57 & 7.8 / 7.7 / 8.2 & 31.7 / 31.7 / 32.2 \\
\bottomrule

\end{tabular}
}
\end{table}

\end{document}